\documentclass[10pt,twocolumn,letterpaper]{article}
\usepackage[hmargin=0.8in,vmargin=0.85in,columnsep=0.3in]{geometry}
\usepackage[table]{xcolor}
\usepackage[numbers,sort&compress]{natbib}
\usepackage[hyphens]{url}
\usepackage{times,amsmath,amssymb,booktabs,tabularx,array,multirow,tikz,algorithm,algpseudocode,microtype,placeins}
\usepackage{graphicx,caption,enumitem,needspace}
\usepackage[skins]{tcolorbox}
\definecolor{abstractbg}{HTML}{EFF2F6}
\definecolor{abstractnote}{HTML}{35465B}
\newtcolorbox{arxivabstract}{%
  enhanced, colback=abstractbg, colframe=abstractbg, boxrule=0pt, arc=8pt,
  shadow={2pt}{-2pt}{0pt}{black!7},
  boxsep=0pt, left=16pt, right=16pt, top=11pt, bottom=11pt,
  before skip=0pt, after skip=0pt
}
\DeclareCaptionJustification{nearfull}{%
  \leftskip=0pt\rightskip=0pt\parfillskip=0pt plus 0.18\linewidth}
\newcommand{\marknum}[1]{\tikz[baseline=(n.base)]\node[circle,fill=black,text=white,inner sep=1pt,font=\scriptsize\bfseries](n){#1};}
\algrenewcommand\alglinenumber[1]{\scriptsize #1}

\makeatletter
\renewcommand{\fnum@algorithm}{\fname@algorithm~\thealgorithm}
\makeatother
\floatstyle{ruled}\restylefloat{algorithm}
\definecolor{harnessaccent}{HTML}{176B7A}
\definecolor{modelaccent}{HTML}{72519B}
\DeclareRobustCommand{\algname}[2]{{%
  \setlength{\fboxsep}{1.2pt}%
  \colorbox{#1!14}{\textcolor{#1}{%
    \fontencoding{OT1}\fontfamily{ComicNeue-TLF}\fontseries{b}\selectfont #2}}}}

\usepackage{pifont,xspace}
\DeclareRobustCommand{\method}{TeachRSI\textsuperscript{2}\xspace}
\newcommand{\fullpar}[2][0]{\par\begingroup
  \setlength{\parfillskip}{0pt plus 0.18\linewidth}%
  \looseness=#1\relax #2\par\endgroup}

\definecolor{linkblue}{RGB}{35,86,140}
\usepackage[colorlinks,allcolors=linkblue]{hyperref}
\hypersetup{pdftitle={TeachRSI2: Learning Structured Visual Relations for Multimodal Agent Self-Improvement}}
\makeatletter
\providecommand{\theHALG@line}{}
\renewcommand{\theHALG@line}{\thealgorithm.\arabic{ALG@line}}
\makeatother

\DeclareRobustCommand{\teachrsilogo}{%
  \tikz[baseline=0.5ex, x=1ex, y=1ex, line cap=round, line join=round]{%
    \filldraw[fill=linkblue!8, draw=linkblue, line width=0.75pt]
      (0.12,0.08) -- (1.98,0.08) -- (1.98,1.86) --
      (1.49,2.35) -- (0.12,2.35) -- cycle;
    \draw[linkblue, line width=0.55pt]
      (1.49,2.35) -- (1.49,1.86) -- (1.98,1.86)
      (0.48,1.55) -- (1.28,1.55)
      (0.48,1.19) -- (1.28,1.19)
      (0.48,0.83) -- (1.05,0.83);
    \filldraw[fill=white, draw=linkblue, line width=0.75pt]
      (1.02,0.49) -- (1.23,1.11) -- (2.36,2.24) --
      (2.77,1.83) -- (1.64,0.70) -- cycle;
    \draw[linkblue, line width=0.55pt]
      (1.23,1.11) -- (1.64,0.70)
      (2.16,2.04) -- (2.57,1.63);
  }%
}
\title{\bfseries \texorpdfstring{\teachrsilogo\enspace}{}%
  TeachRSI\textsuperscript{2}: Learning Structured Visual Relations\\
  for Multimodal Agent Self-Improvement}
\author{%
  Ke He\textsuperscript{1,*}\quad
  Yonghan Peng\textsuperscript{1,*}\quad
  Chenxu Zhang\textsuperscript{2,*}\quad
  Xiang Zhang\textsuperscript{1}\quad
  ZePing Tian\textsuperscript{3}\\
  Yuejia Li\textsuperscript{5}\quad
  Zihan Tan\textsuperscript{1}\quad
  Qishun Yang\textsuperscript{4}\quad
  Bin Yang\textsuperscript{1,\textdagger}\\[0.65em]
  {\small \textsuperscript{1}Wuhan University\quad
  \textsuperscript{2}Imperial College London\quad
  \textsuperscript{3}The Chinese University of Hong Kong Shenzhen}\\
  {\small \textsuperscript{4}King Abdullah University of Science and Technology\quad
  \textsuperscript{5}University of California, Berkeley}%
}
\date{}

\makeatletter
\renewcommand{\@maketitle}{%
  \begin{center}
    {\LARGE \@title\par}
    \vspace{0.8em}
    {\large
      \begin{tabular}[t]{c}
        \@author
      \end{tabular}\par}
  \end{center}
  \vspace{0.6em}
}
\makeatother

\definecolor{tblhead}{RGB}{235,235,235}
\definecolor{tblstripe}{RGB}{245,245,247}
\definecolor{ours}{RGB}{252,232,232}
\newcommand{\resultcell}{\textemdash}
\newcommand{\yes}{\ensuremath{\checkmark}}

\newcolumntype{Y}{>{\centering\arraybackslash}X}
\newcolumntype{L}[1]{>{\raggedright\arraybackslash}p{#1}}
\newcommand{\tablelook}{%
  \hfuzz=0.5pt%
  \setlength{\tabcolsep}{3pt}%
  \setlength{\arrayrulewidth}{0.45pt}%
  \setlength{\doublerulesep}{1.0pt}%
  \setlength{\aboverulesep}{0pt}%
  \setlength{\belowrulesep}{0pt}%
  \renewcommand{\arraystretch}{1.17}%
  \arrayrulecolor{black!75}%
}
\newcommand{\thcell}[1]{\shortstack[c]{#1}}
\newcommand{\modelcell}[1]{%
  \cellcolor{white}\multirow{-11}{=}{\raggedright\bfseries\mbox{#1}}}

\newcommand{\scorepair}[2]{{\renewcommand{\arraystretch}{1.0}\begin{tabular}[c]{@{}c@{}}#1\,/\,\\#2\end{tabular}}}
\newcommand{\uncert}[2]{\mbox{#1\,$\pm$\,#2}}

\begin{document}
\twocolumn[{%
\maketitle
\begin{arxivabstract}
\noindent Recursive self-improvement (RSI) of multimodal agents requires turning visual interaction experience into reusable capabilities. Existing RSI methods reuse interaction experience by refining external instructions, programs, and triggering rules while keeping model parameters fixed. Two challenges in this process motivate our approach: 1) Continued accumulation of external guidance can increase execution overhead and dependence on specialized rules, motivating the internalization of reusable capabilities into model parameters. 2) Small targets, cluttered backgrounds, and unclear evidence can make raw visual experience difficult to learn from directly, motivating more learnable training examples. We propose \textbf{Teach}er-guided dual-loop \textbf{R}ecursive \textbf{S}elf-\textbf{I}mprovement (TeachRSI$^2$), which couples harness evolution with curriculum-based model learning. \textbf{Harness-RSI} uses teacher feedback to refine execution strategies and identify model weaknesses. \textbf{Model-RSI} constructs multiple views of the same task through image augmentation while preserving necessary evidence, and organizes the resulting examples into an easy-to-hard curriculum, progressing from clearer views to more challenging visual conditions. Alongside verified answer and action supervision, a visual relation loss encourages consistent predictions under answer-preserving transformations and correct answer mappings under reference-frame changes. Harness and model updates alternate, with the refined harness supplying training experience and the updated model enabling further harness refinement and the removal of redundant guidance when performance is preserved. Experiments on TIR-Bench, MME-RealWorld-Lite, and BLINK demonstrate consistent improvements over harness, model, and joint optimization baselines across student backbones, supporting the effectiveness of learning from structured visual experience. \textit{The source code will be released.}
\par\smallskip
\noindent{\small\itshape\color{abstractnote}%
  \textsuperscript{*}~Equal contribution.\qquad
  \textsuperscript{\textdagger}~Corresponding author.}
\end{arxivabstract}
\vspace{0.8em}
}]

\section{Introduction}
\label{sec:intro}

\fullpar{Multimodal agents acquire visual evidence through tool use~\cite{yao2023react} and make decisions from the resulting observations~\cite{gupta2023visprog,suris2023vipergpt,hu2024sketchpad}. Their successes and failures provide experience for recursive self-improvement (RSI)~\cite{shinn2023reflexion,madaan2023selfrefine,zhao2024expel}. Existing methods reuse such experience by refining skills and instructions~\cite{jiang2026xskill,agrawal2025gepa,wang2024voyager} within the harness, the software that organizes an agent's tools and execution~\cite{khattab2024dspy}. Agents can thus adapt their behavior through external guidance while keeping the underlying model parameters fixed.}

\fullpar{As experience accumulates, however, external guidance can become longer and more specialized, increasing execution overhead and dependence on task-specific rules~\cite{sun2026herd,ye2026harnesszero}. Model learning offers a way to internalize reusable capabilities~\cite{liu2023llava,li2023blip2,hsieh2023distilling}, and recent work has begun coupling harness evolution with parameter updates~\cite{sun2026herd,shang2026rlharness}. For multimodal agents, a central challenge is to turn harness experience into transferable visual skills that can support further improvement with less external guidance.}

\fullpar{Raw visual experience is not always easy to learn from. Small targets and cluttered backgrounds can obscure the evidence needed to solve a task~\cite{wu2024vstar,zhang2025mmerealworld,fu2024blink}. A crop that makes small text legible, for example, can provide an easier training view before the model learns from the full image. This motivates constructing multiple views of the same task~\cite{cubuk2019autoaugment,cubuk2020randaugment,gong2021keepaugment} and organizing them into an easy-to-hard curriculum~\cite{bengio2009curriculum,kumar2010selfpaced,wang2022curriculumsurvey}. Learning across these views also requires respecting their semantics~\cite{hendrycks2020augmix,xie2020uda,sohn2020fixmatch}. Recognizing the same text should preserve the answer, whereas measuring an object's area relative to a crop or the full image requires different answers linked by a known conversion. Training should therefore account for both visual difficulty and the answer relations induced by each operation.}

\fullpar[-1]{We propose Teacher-guided dual-loop Recursive Self-Improvement (TeachRSI$^2$). Harness-RSI uses teacher feedback to improve execution strategies and identify model weaknesses. Model-RSI generates and verifies augmented views, orders training examples by difficulty, and combines answer and action supervision with a visual relation loss. Harness and model updates alternate, allowing new capabilities to inform subsequent harness refinement and performance-preserving removal of redundant guidance. Experiments on TIR-Bench~\cite{li2026tirbench}, MME-RealWorld-Lite~\cite{zhang2025mmerealworld}, and BLINK~\cite{fu2024blink} demonstrate gains across four students from two model families.}

\Needspace{4\baselineskip}
\noindent\textbf{Contributions.}
\begin{enumerate}[label=\protect\marknum{\arabic*},leftmargin=*,itemsep=2pt,topsep=3pt]
\item \fullpar{We introduce TeachRSI$^2$, a dual-loop framework that alternates harness refinement with model learning to internalize external interaction experience.}
\item \fullpar{We construct training views of varying difficulty from the same task through image augmentation and organize verified examples into an easy-to-hard curriculum.}
\item \fullpar{We introduce visual relation supervision alongside answer and action targets, enforcing prediction consistency for answer-preserving transformations and correct answer mappings for reference-frame changes.}
\item \fullpar{We evaluate TeachRSI$^2$ on three benchmarks with four student backbones and use component ablations to assess harness refinement, model learning, curriculum ordering, relation supervision, and compaction.}
\end{enumerate}

\section{Related Work}
\label{sec:related}
\subsection{Multimodal Harness Evolution}
\fullpar{One line of agent self-improvement changes external memory and execution procedures while leaving model parameters fixed~\cite{shinn2023reflexion,zhao2024expel,wang2024voyager}. Agent Workflow Memory~\cite{wang2024awm} extracts reusable workflows, Dynamic Cheatsheet~\cite{suzgun2025dc} accumulates task strategies, and XSkill~\cite{jiang2026xskill} organizes multimodal experiences into reusable skills. A complementary line directly searches the harness~\cite{hu2025adas,zhang2025aflow,khattab2024dspy,yuksekgonul2025textgrad}: GEPA~\cite{agrawal2025gepa} and SkillOpt~\cite{yang2026skillopt} optimize textual guidance, whereas Video-RSI~\cite{luo2026videorsi} evolves executable visual procedures. These approaches make past executions reusable, but an improved external procedure does not specify which capabilities should be internalized by the model. \method retains harness refinement while turning the visual operations and returned observations inside its executions into answer, action, and relation supervision for model learning.}

\subsection{Harness-to-Model Transfer}
\fullpar{A second line transfers externally supported behavior into model parameters~\cite{hsieh2023distilling,schick2023toolformer}. Visual Program Distillation~\cite{hu2024vpd} distills tools and programmatic reasoning into vision-language models, while Harness-Zero~\cite{ye2026harnesszero} and Harness-Aware Distillation~\cite{choi2026had} derive training signals from harness-guided executions. Complementary to demonstration-based transfer, Radioactive Feedback~\cite{li2026radioactive} studies how evaluator-specific behavioral preferences can be transferred and traced through label-only rewards. Joint optimization methods further connect external and parametric improvement. BetterTogether~\cite{soylu2024bettertogether} and MetaTuner~\cite{bo2026metatuner} combine prompt and parameter optimization, whereas HERD~\cite{sun2026herd} and RLHarness~\cite{shang2026rlharness} couple evolving procedures with distillation or reinforcement learning. These methods establish that external guidance and parameter learning can reinforce each other, but they do not explicitly model how a visual operation changes the meaning or reference frame of an answer. \method addresses this gap with verified operation-grounded answer relations and fixed-counterpart acceptance tests for harness and model updates.}

\subsection{Data Generation and Curriculum}
\fullpar{Data generation converts experience into examples~\cite{wang2023selfinstruct,kim2024multimodalselfinstruct}, while curricula organize their presentation~\cite{kumar2010selfpaced,wang2022curriculumsurvey}. STaR~\cite{zelikman2022star} iteratively generates reasoning traces and trains on those yielding correct answers, allowing an updated model to produce subsequent training data. In vision, AutoAugment~\cite{cubuk2019autoaugment} searches for image transformation policies that improve classification performance. Curriculum learning~\cite{bengio2009curriculum} organizes training from easier examples toward more difficult ones. These ideas motivate our use of visual interactions to construct a learning progression for multimodal agents. Given a source task, \method generates augmented views that preserve necessary evidence, verifies their targets, and organizes them by difficulty.}

\fullpar{Related visual generation work also explicitly models structure. See Before You Code~\cite{li2026seebeforeyoucode} combines visual planning with structured post-render diagnostics for educational animation generation. OneWorld~\cite{he2026oneworld} encourages consistency of the underlying physical mechanism across action-conditioned futures while preserving action-dependent outcomes. These approaches address spatial quality and physical consistency in generation, respectively, whereas \method learns operation-grounded answer relations across augmented observations, preserving predictions when appropriate and mapping them when the reference frame changes.}

\begin{figure}[!t]
\centering
\includegraphics[width=\linewidth,pagebox=cropbox,clip]{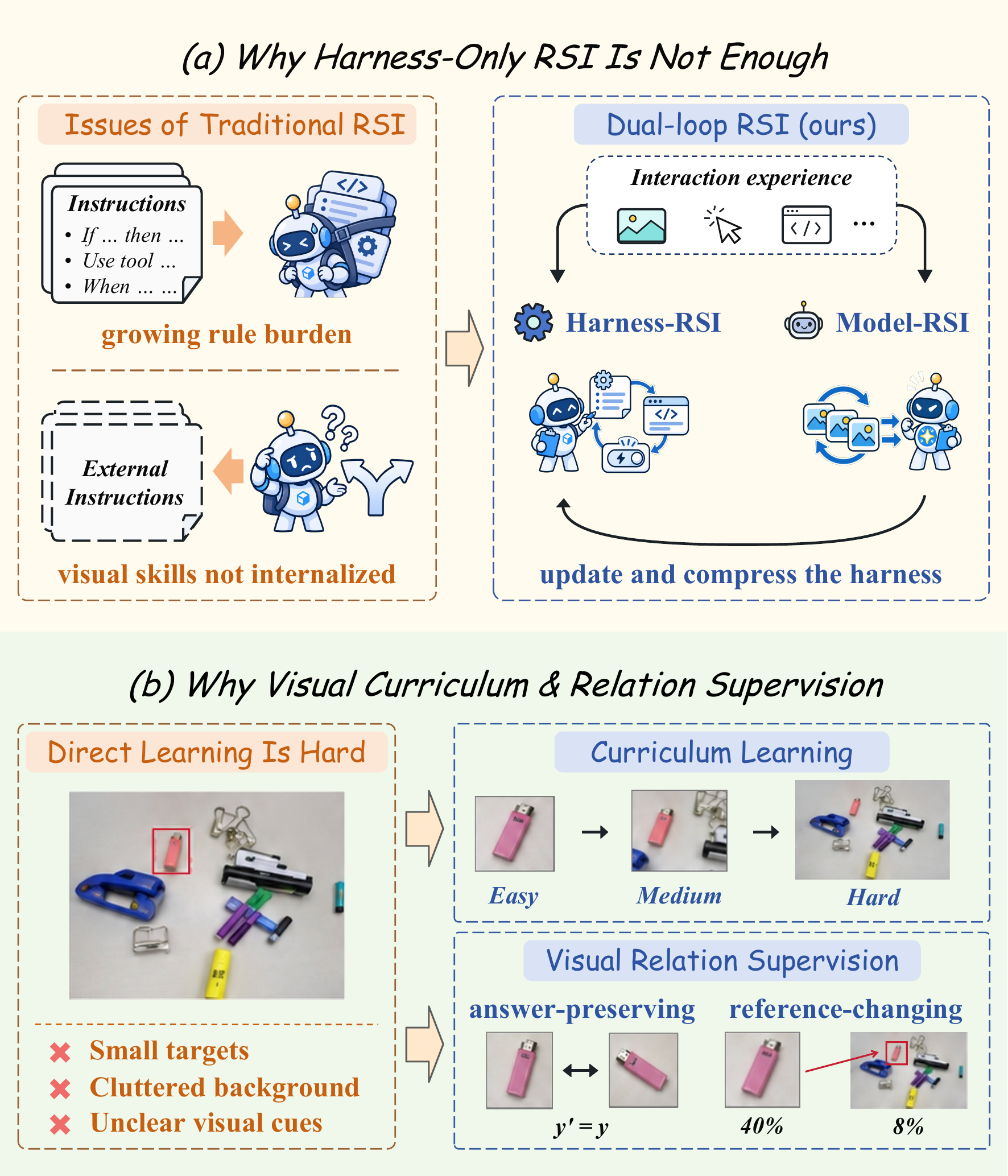}
\caption{Motivation for \method. (a) Harness refinement and model learning jointly reuse visual interaction experience, reducing reliance on accumulating task-specific rules. (b) A curriculum organizes views from easy to hard, while visual relation learning connects answer-preserving transformations and reference-frame changes, including crop-to-image area conversion.}
\label{fig:motivation}
\end{figure}

\section{Method}
\label{sec:methods}

\begin{figure*}[t]
\centering
\includegraphics[width=\textwidth]{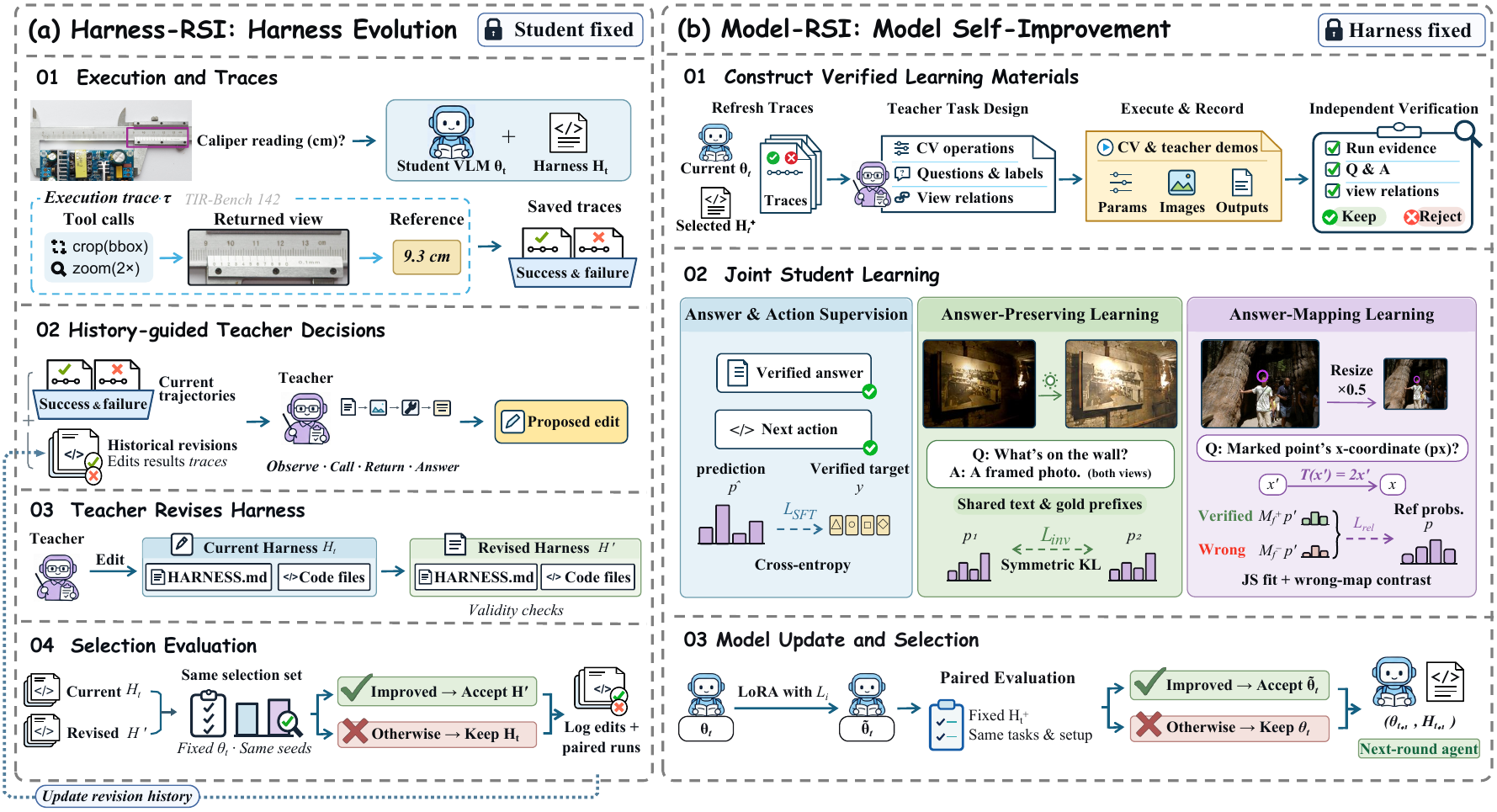}
\caption{Overview of \method. Harness-RSI refines programs and instructions with the model fixed. Model-RSI uses verified visual experience to construct training views, organize a curriculum, and learn answers, actions, and visual relations with the harness fixed. Accepted model updates support subsequent harness refinement and consolidation.}
\label{fig:architecture}
\end{figure*}

\subsection{Formulation}
\label{sec:problem_formulation}
\fullpar{\noindent\textbf{Agent state and interaction.} A visual task is an image--question pair $x=(I,q)$. The agent state $S_t=(\theta_t,H_t)$ contains model parameters $\theta_t$ and a harness $H_t$ comprising operating instructions, reusable programs, and tool-invocation rules. Given $x$, the model acts through the harness to produce a trajectory $\tau=(x,h,\hat y)$, where $h$ records model outputs, tool arguments, and actual returns, and $\hat y$ is the final answer. Source images, transformation parameters, and component versions accompany each record. This provenance links predictions to their evidence and execution procedures, distinguishing incorrect answers from misinterpreted tool results.}

\fullpar{\noindent\textbf{Optimization objective.} Let $\mathcal E$ be a fixed selection set of task--seed pairs, and let $R_{\theta,H}(x,s)\in\{0,1\}$ denote final-answer correctness under seed $s$. We evaluate a state by}
\begin{equation}
A(\theta,H)=\frac{1}{|\mathcal E|}\sum_{(x,s)\in\mathcal E}R_{\theta,H}(x,s).
\label{eq:selection_score}
\end{equation}
\fullpar{The goal is to improve $A$ by updating external procedures and model parameters within fixed search and training budgets. Harness-RSI first obtains $H_t^+$ with $\theta_t$ fixed. Its executions supply the verified views and relations used by Model-RSI to learn $\theta_t^+$ with $H_t^+$ fixed. Optional consolidation then removes redundant guidance if correctness is preserved. Training interactions provide proposals and supervision, selection tasks govern acceptance under matched execution settings, and held-out test tasks measure final performance. Figure~\ref{fig:architecture} summarizes this division of responsibilities.}

\subsection{Harness Refinement}
\label{sec:harness_refinement}
\fullpar{\noindent\textbf{Trace-based proposal generation.} With $\theta_t$ frozen, the teacher examines successful and failed training executions together with the current harness and search history. Successful traces identify reusable procedures; failures expose missing operations, incorrect tool arguments, or misinterpretation of returned evidence. The teacher turns these observations into edits to instructions or executable programs. For example, copying a crop-relative fraction into a global answer motivates guidance that records the crop footprint and requested denominator. The archive stores parent revisions, candidates, and outcomes for subsequent proposals (Algorithm~\ref{alg:harness}).}

\fullpar{\noindent\textbf{Execution and acceptance.} At each search step, at most $B$ candidates are proposed from the current incumbent $H$. Candidates run in isolated workspaces under the same underlying tool implementations, model, and execution limits. Executable candidates are assessed using Eq.~\ref{eq:selection_score}; the resulting selection gain is}
\begin{equation}
\Delta_H(\widetilde H;H)=A(\theta_t,\widetilde H)-A(\theta_t,H).
\label{eq:harness_gain}
\end{equation}
\fullpar{A candidate replaces the incumbent only when its selection gain is positive; otherwise the incumbent is retained. Failed execution does not yield an accepted revision, and invalid or duplicate proposals are recorded in search accounting. Repeating proposal, execution, and comparison within the allocated budget yields $H_t^+$. Because the model is unchanged, an accepted gain is attributable to the modified external procedure on the shared selection set.}

\fullpar{\noindent\textbf{Transfer to model learning.} The accepted harness collects fresh training executions under $(\theta_t,H_t^+)$. These records provide source tasks, observed weaknesses, and tool-generated views for the curriculum. The next section describes how these records support training-view construction, target verification, and curriculum design.}

\begin{algorithm}[t]
\caption{\algname{harnessaccent}{Harness-RSI}: Harness refinement}
\label{alg:harness}
\small
\begin{algorithmic}[1]
\Require Fixed $\theta_t$, initial $H_t$, training tasks $\mathcal D$
\Require Selection pairs $\mathcal E$, search steps $J$, limit $B$
\Ensure Accepted $H_t^+$ and fresh training traces $\mathcal T_t^+$
\State $H\gets H_t$; initialize the revision archive
\For{$j=1,\ldots,J$}
  \State Collect training traces under $(\theta_t,H)$
  \State Teacher proposes up to $B$ harness edits
  \State Execute in isolation; archive edits and outcomes
  \For{each valid, novel candidate $\widetilde H$}
    \If{$\Delta_H(\widetilde H;H)>0$}
      \State $H\gets\widetilde H$
    \EndIf
  \EndFor
\EndFor
\State $H_t^+\gets H$; collect $\mathcal T_t^+$ on $\mathcal D$
\State \Return $H_t^+,\mathcal T_t^+$
\end{algorithmic}
\end{algorithm}

\subsection{Visual Curriculum}
\label{sec:visual_curriculum}
\label{sec:verified_supervision}
\begin{figure}[t]
\centering
\includegraphics[width=\linewidth,pagebox=cropbox,clip]{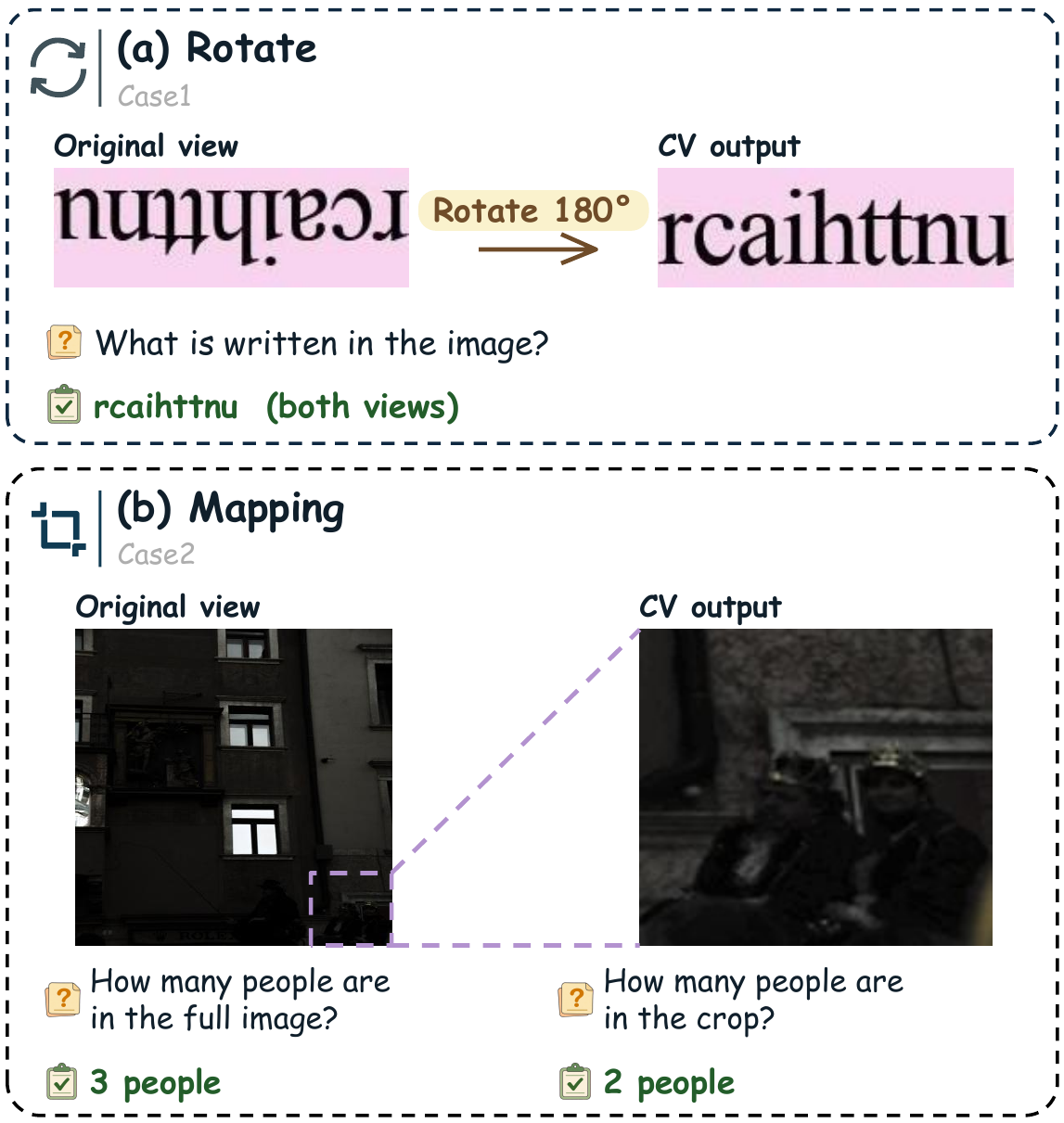}
\caption{Training-view examples. (a) A $180^\circ$ rotation preserves the text-recognition target. (b) Cropping changes the counting scope: the full image contains three people, while the crop contains two. The count relation requires verified evidence outside the crop; visual difficulty is assessed separately.}
\label{fig:training_examples}
\end{figure}

\fullpar{\noindent\textbf{View construction and verification.} For each source task $i$, the teacher uses its image, question, and execution history to construct training views through actual image operations. Cropping, resizing, and contrast adjustment can expose small targets or reduce distracting context while preserving the evidence required by the question. Each view $v_{ij}$ records its image, question, verified target, and operation provenance. Views of the same task may differ in visual difficulty without changing the requested answer. If an operation changes the quantity or reference frame being queried, the corresponding question and target are recorded explicitly rather than inheriting the original label. The teacher checks answer targets against task references or independently reviewed visual content. Views lacking sufficient evidence are excluded. Verified views form a direct-answer set $\mathcal D_i$, while exact execution prefixes paired with verified next tool calls or final answers form an action set $\mathcal A_i$. A view pair receives additional relation supervision only when its operation is traceable, both answers are supported, and the proposed relation has an independent basis in the recorded transformation and visual evidence. Pairs with unresolved relations retain only their individually verified targets. Figure~\ref{fig:training_examples} illustrates answer-preserving rotation and a crop that changes the counting scope.}

\fullpar{\noindent\textbf{Progressive admission.} Before optimization, the teacher assigns each view an ordinal difficulty $d_{ij}$ from target visibility, distracting context, and required reasoning, rather than transformation severity alone. Stage $s$ admits $\mathcal V_i^{(s)}=\{j:d_{ij}\le b_s\}$ with nondecreasing thresholds $b_s\le b_{s+1}$. Clearer views precede harder ones, including the original image when appropriate. Targets and difficulty labels remain fixed within a round, and earlier views stay eligible. A source becomes active with its first eligible view; paired losses activate only when both endpoints are admitted. Verified action examples accompany the views. This schedule controls when verified supervision becomes available without changing the meaning of its targets.}

\subsection{Model Learning}
\label{sec:model_learning}
\label{sec:visual_relation}
\fullpar{\noindent\textbf{Answer and action supervision.} With $H_t^+$ fixed, Model-RSI learns from verified answers, actions, and relations among eligible views. For context $c$ and target sequence $y$, we define the token-averaged cross-entropy loss as}
\begin{equation}
L_{\mathrm{CE}}(c,y)=-\frac{1}{|y|}\sum_{k=1}^{|y|}\log\pi_\theta(y_k\mid c,y_{<k}).
\label{eq:cross_entropy}
\end{equation}
\fullpar{Contexts contain images, questions, harness instructions, and history; only target assistant tokens receive labels. Let $L_D^i$ and $L_A^i$ average the answer and action losses for source $i$. Their supervised objective is}
\begin{equation}
L_{\mathrm{SFT}}^i=\frac{L_D^i+\omega_i L_A^i}{1+\omega_i}.
\label{eq:supervised_objective}
\end{equation}
\fullpar{Here $\omega_i=\omega\ge0$ when action targets exist and zero otherwise. Group averaging prevents sources with more views or actions from dominating through their size.}

\fullpar{\noindent\textbf{Answer-preserving consistency.} Paired contexts share a verified answer, identical non-image content, and aligned target tokens. Let $p_k$ and $p'_k$ be next-token distributions conditioned on the same answer prefix. The symmetric divergence between the distributions is}
\begin{equation}
D_{\mathrm{SKL}}(p,q)=\frac{D_{\mathrm{KL}}(p\Vert q)+D_{\mathrm{KL}}(q\Vert p)}{2}.
\label{eq:symmetric_kl}
\end{equation}
\Needspace{4\baselineskip}
\fullpar{For pair $e$, the $n$-token consistency loss is~\cite{liang2021rdrop}}
\begin{equation}
L_{\mathrm{inv}}^e=\frac{1}{n}\sum_{k=1}^{n}D_{\mathrm{SKL}}(p_k,p'_k).
\label{eq:invariance}
\end{equation}

\fullpar{\noindent\textbf{Reference-changing relations.} For a reference-changing pair $e$, a verified mapping $f_e^\star$ connects local and global answer frames. If an object is fully contained in crop $C_e$ of image $I_e$, its area fractions satisfy $r_g=\alpha_e r_l$, where $\alpha_e=|C_e|/|I_e|$. Areas are measured in source coordinates, preserving the conversion after resizing. Coordinate mappings use recorded offsets and scales, and compositions require compatible intermediate domains. These conditions establish a relation independently of simply fitting the two answer labels. To compare semantic answers, each context has a fixed candidate set with canonical strings $s(a)$. A softmax over scores $u_\theta(a,c)=-L_{\mathrm{CE}}(c,s(a))$ gives a candidate distribution. A mapping $f$ defines a fixed nonnegative transport matrix $M_f$ whose columns sum to one. Exact matches transfer mass directly, while intermediate values use predefined interpolation within the covered range. For the candidate distributions $p_\theta^l$ and $p_\theta^g$, the mapping discrepancy is}
\begin{equation}
d_e(f)=D_{\mathrm{JS}}(p_\theta^g\Vert M_f p_\theta^l).
\label{eq:mapping_discrepancy}
\end{equation}
\fullpar{Wrong mappings $\mathcal F_e^-$ represent errors such as omitting an offset or copying a local fraction into a global answer. They use the same candidate domains and differ from the verified mapping at the target beyond numerical and discretization tolerances. Writing $d_e^+=d_e(f_e^\star)$ and $\delta_e(f)=d_e^+-d_e(f)$, we combine relation fitting with wrong-map contrast}
\begin{equation}
L_{\mathrm{rel}}^e=d_e^++\gamma\log\!\left(1+\sum_{f\in\mathcal F_e^-}e^{\delta_e(f)/\beta}\right).
\label{eq:relation_objective}
\end{equation}
\fullpar{Here $\gamma\ge0$ weights the contrast and $\beta>0$ is its temperature. Both branches receive gradients and answer supervision, while targets, candidates, and mappings remain fixed. Therefore, even when mutually consistent answers are incorrect, our method uses answer supervision to guide each view's prediction toward its verified target.}

\fullpar{\noindent\textbf{Joint optimization.} For source $i$, let $L_{\mathrm{inv}}^i$ and $L_{\mathrm{rel}}^i$ average the pair losses $L_{\mathrm{inv}}^e$ and $L_{\mathrm{rel}}^e$ over eligible invariant and reference-changing pairs, respectively. The stage objective is}
\begin{equation}
L_i=L_{\mathrm{SFT}}^i+\lambda_{\mathrm{inv}}L_{\mathrm{inv}}^i+\lambda_{\mathrm{rel}}L_{\mathrm{rel}}^i.
\label{eq:source_objective}
\end{equation}
\fullpar{Empty pair sets contribute zero; both relation weights are nonnegative. Model-RSI minimizes the batch mean using AdamW~\cite{loshchilov2019adamw} to update LoRA adapters~\cite{hu2022lora} in attention query, key, value, and output projections, with pretrained weights and the harness fixed. Algorithm~\ref{alg:training} uses a verified snapshot $\mathcal Z$ of views, action targets, relations, and difficulty labels. Its schedule $\mathcal C=\{(b_s,U_s)\}_{s=1}^{K}$ specifies thresholds and optimizer steps per stage; $\phi$ denotes the candidate model's trainable LoRA parameters.}

\begin{algorithm}[t]
\caption{\algname{modelaccent}{Model-RSI}: Curriculum learning}
\label{alg:training}
\small
\begin{algorithmic}[1]
\Require $\theta_t$, fixed $H_t^+$, verified data $\mathcal Z$, schedule $\mathcal C$
\Ensure Updated model candidate $\widetilde\theta$
\State Initialize $\widetilde\theta\gets\theta_t$ with LoRA parameters $\phi$
\State Freeze harness, base weights, and supervision
\For{$s=1,\ldots,K$}
  \State Admit views with $d_{ij}\le b_s$
  \State Retain pairs with both views admitted
  \For{$u=1,\ldots,U_s$}
    \State Sample sources $\mathcal B$ and verified targets
    \State Compute $L_i$ using Eq.~\ref{eq:source_objective}
    \State $L_{\mathrm{batch}}\gets |\mathcal B|^{-1}\sum_{i\in\mathcal B}L_i$
    \State $\phi\gets\operatorname{AdamW}(\phi,\nabla_{\phi}L_{\mathrm{batch}})$
  \EndFor
\EndFor
\State \Return $\widetilde\theta$
\end{algorithmic}
\end{algorithm}

\subsection{Dual-Loop Optimization}
\label{sec:alternating}
\fullpar{\noindent\textbf{Model acceptance.} Each round freezes a verified training snapshot generated under $(\theta_t,H_t^+)$. Eligible historical records may supplement it with their original contexts, while labels and relation mappings remain fixed throughout optimization. Algorithm~\ref{alg:training} returns a candidate $\widetilde\theta$, whose selection gain is measured with the same harness on both sides:}
\begin{equation}
\Delta_\theta(\widetilde\theta;\theta_t)=A(\widetilde\theta,H_t^+)-A(\theta_t,H_t^+).
\label{eq:model_gain}
\end{equation}
\fullpar{The candidate becomes $\theta_t^+$ only if $\Delta_\theta>0$; otherwise $\theta_t^+=\theta_t$. Equations~\ref{eq:harness_gain} and~\ref{eq:model_gain} use the same correctness measure while fixing complementary components. This isolates each component's gain and retains harness improvements when model updates are rejected.}

\fullpar{\noindent\textbf{Consolidation and recurrence.} After an accepted model update, the teacher may propose a shorter harness $\widehat H$ by merging repeated guidance while preserving executable programs. Let $T(H)$ count harness tokens with a fixed tokenizer. Consolidation is accepted only when $A(\theta_t^+,\widehat H)\ge A(\theta_t^+,H_t^+)$ and $T(\widehat H)<T(H_t^+)$. The next state uses $\theta_{t+1}=\theta_t^+$ and the retained harness as $H_{t+1}$. New interactions are then collected under this state, so later harness proposals respond to the updated model's strengths and remaining errors. Acceptance preserves the observed selection score at each transition; it does not guarantee test-set improvement, which is evaluated separately on held-out tasks.}

\begin{table*}[!t]
\centering
\caption{End-to-end comparison from identical initial students in two VLM families. Results report Average@4 (Avg.@4) and Pass@4 in percent ($\uparrow$), with equal benchmark weights for Macro Avg. Methods are grouped by harness, model, or joint (H+M) updates.}
\label{tab:main}
\begingroup
\newcommand{\qwenicon}{\IfFileExists{fig/icons/qwen_official.png}{\raisebox{-0.45ex}{\includegraphics[height=2.3ex,trim=24bp 21.6bp 319.2bp 27.12bp,clip]{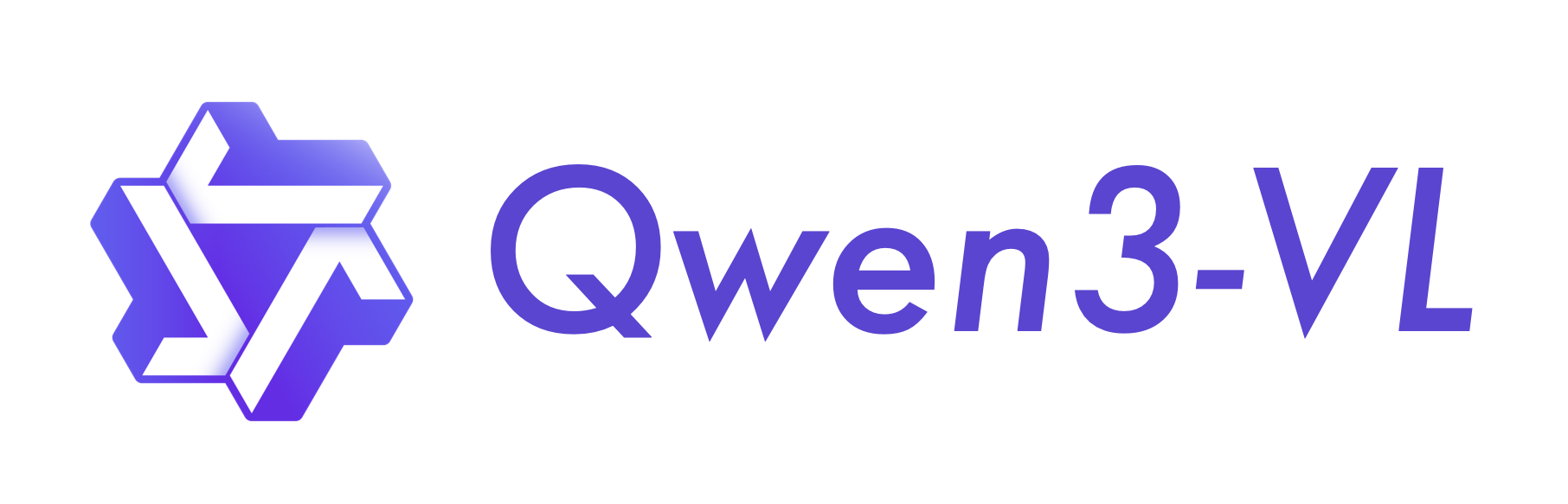}}\hspace{0.4em}}{}}
\newcommand{\internvlicon}{\IfFileExists{fig/icons/internvl_official.png}{\raisebox{-0.45ex}{\includegraphics[height=2.3ex,trim=42bp 20bp 710bp 20bp,clip]{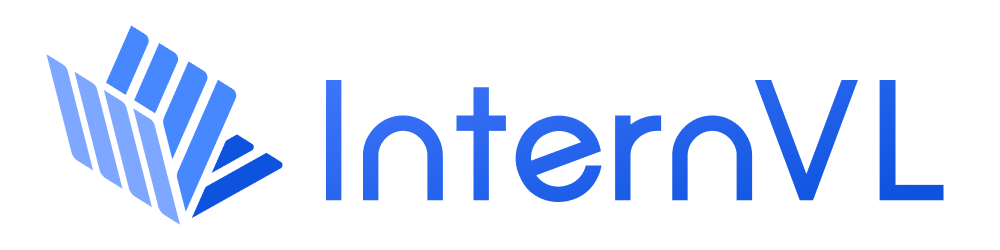}}\hspace{0.4em}}{}}
\tablelook
\definecolor{tblhead}{RGB}{238,239,241}
\definecolor{tblstripe}{RGB}{248,248,249}
\definecolor{ours}{RGB}{252,238,238}
\renewcommand{\resultcell}{\textemdash}
\renewcommand{\modelcell}[1]{\cellcolor{white}\multirow{-11}{=}{\centering\bfseries #1}}
\fontsize{8}{8.5}\selectfont
\renewcommand{\arraystretch}{1.12}
\setlength{\tabcolsep}{1.6pt}
\begin{tabularx}{0.96\textwidth}{>{\centering\arraybackslash}p{0.04025\textwidth}|L{0.1495\textwidth}|>{\centering\arraybackslash}p{0.09775\textwidth}||YY|YY|YY|YY}
\toprule[0.8pt]
\rowcolor{tblhead}
& & &
\multicolumn{2}{c|}{{\fontsize{8}{9.3}\selectfont\textbf{TIR-Bench}}} &
\multicolumn{2}{c|}{{\fontsize{8}{9.3}\selectfont\textbf{MME-RealWorld-Lite}}} &
\multicolumn{2}{c|}{{\fontsize{8}{9.3}\selectfont\textbf{BLINK}}} &
\multicolumn{2}{c}{{\fontsize{8}{9.3}\selectfont\textbf{Macro Avg.}}}\\
\rowcolor{tblhead}
\multirow{-2}{*}{\textbf{Type}} & \multirow{-2}{*}{\textbf{Method}} & \multirow{-2}{*}{\textbf{Venue}} & Avg.@4 & Pass@4 & Avg.@4 & Pass@4 & Avg.@4 & Pass@4 & Avg.@4 & Pass@4\\
\midrule[0.45pt]
\rowcolor{tblhead}\multicolumn{11}{l}{\rule{0pt}{10pt}\qwenicon \textbf{Qwen3-VL-2B}}\\
\midrule[0.35pt]
\rowcolor{tblstripe} \cellcolor{white} & AWM~\cite{wang2024awm} & {\fontsize{7.1}{8.4}\selectfont ICML-25} & 16.45 & 29.93 & 42.04 & 61.69 & 52.42 & 80.84 & 36.97 & 57.49 \\
\rowcolor{white} \cellcolor{white} & DC~\cite{suzgun2025dc} & {\fontsize{7.1}{8.4}\selectfont EACL-26} & 16.12 & 28.62 & 41.43 & 61.92 & 45.58 & 73.26 & 34.37 & 54.60 \\
\rowcolor{white} \cellcolor{white} & XSkill~\cite{jiang2026xskill} & {\fontsize{7.1}{8.4}\selectfont ICML-26} & 17.60 & 32.57 & 45.94 & 65.92 & 51.89 & 76.63 & 38.48 & 58.37 \\
\rowcolor{tblstripe} \cellcolor{white} & GEPA~\cite{agrawal2025gepa} & {\fontsize{7.1}{8.4}\selectfont ICLR-26} & 15.13 & 31.25 & 48.55 & 60.80 & 55.05 & 79.58 & 39.58 & 57.21 \\
\rowcolor{white} \cellcolor{white} \multirow{-5}{*}{\rotatebox{90}{Harness}} & SkillOpt~\cite{yang2026skillopt} & {\fontsize{7.1}{8.4}\selectfont NeurIPS-26} & 15.13 & 24.67 & 47.72 & 65.70 & 52.37 & 76.63 & 38.41 & 55.67 \\
\midrule[0.45pt]
\rowcolor{white} \cellcolor{white} & VTool-R1$^{*}$~\cite{wu2026vtoolr1} & {\fontsize{7.1}{8.4}\selectfont ICLR-26} & 19.81 & 36.07 & 46.20 & 62.96 & 54.55 & 73.43 & 40.19 & 57.49 \\
\rowcolor{tblstripe} \cellcolor{white}  & AdaReasoner$^{*}$~\cite{song2026adareasoner} & {\fontsize{7.1}{8.4}\selectfont ICLR-26} & 15.57 & 28.42 & 42.04 & 65.56 & 51.92 & 70.98 & 36.51 & 54.98 \\
\rowcolor{white} \cellcolor{white} \multirow{-3}{*}{\rotatebox{90}{Model}} & VisTA$^{*}$~\cite{huang2026vista} & {\fontsize{7.1}{8.4}\selectfont CVPR-26} & 12.43 & 16.94 & 40.56 & 48.89 & 47.73 & 53.50 & 33.57 & 39.78 \\
\midrule[0.45pt]
\rowcolor{tblstripe} \cellcolor{white} & {\fontsize{7.1}{8.4}\selectfont BetterTogether$^{*}$~\cite{soylu2024bettertogether}} & {\fontsize{7.1}{8.4}\selectfont EMNLP-24} & 16.26 & 25.14 & 50.83 & 65.19 & 49.65 & 52.80 & 38.91 & 47.71 \\
\rowcolor{white} \cellcolor{white} & {\fontsize{7.1}{8.4}\selectfont MetaTuner$^{*}$~\cite{bo2026metatuner}} & {\fontsize{7.1}{8.4}\selectfont ICLR-26} & 23.77 & 51.37 & 52.41 & 68.89 & 57.17 & 77.62 & 44.45 & 65.96 \\
\rowcolor{ours} \cellcolor{white} \multirow{-3}{*}{\rotatebox{90}{H+M}} & \textbf{\method} & {\fontsize{7.1}{8.4}\selectfont \textemdash} & \textbf{27.73} & \textbf{56.83} & \textbf{58.15} & \textbf{74.07} & \textbf{63.11} & \textbf{84.27} & \textbf{49.66} & \textbf{71.72} \\
\midrule[0.5pt]
\rowcolor{tblhead}\multicolumn{11}{l}{\rule{0pt}{10pt}\qwenicon \textbf{Qwen3-VL-8B}}\\
\midrule[0.35pt]
\rowcolor{tblstripe} \cellcolor{white} & AWM~\cite{wang2024awm} & {\fontsize{7.1}{8.4}\selectfont ICML-25} & 17.43 & 29.28 & 35.58 & 46.77 & 61.21 & 80.63 & 38.07 & 52.23 \\
\rowcolor{white} \cellcolor{white} & DC~\cite{suzgun2025dc} & {\fontsize{7.1}{8.4}\selectfont EACL-26} & 15.87 & 24.67 & 14.09 & 22.27 & 61.79 & 80.63 & 30.58 & 42.52 \\
\rowcolor{white} \cellcolor{white} & XSkill~\cite{jiang2026xskill} & {\fontsize{7.1}{8.4}\selectfont ICML-26} & 18.01 & 29.61 & 37.69 & 50.11 & 63.58 & 79.58 & 39.76 & 53.10 \\
\rowcolor{tblstripe} \cellcolor{white} & GEPA~\cite{agrawal2025gepa} & {\fontsize{7.1}{8.4}\selectfont ICLR-26} & 17.93 & 26.32 & 54.68 & 64.59 & 64.47 & 74.74 & 45.69 & 55.21 \\
\rowcolor{white} \cellcolor{white} \multirow{-5}{*}{\rotatebox{90}{Harness}} & SkillOpt~\cite{yang2026skillopt} & {\fontsize{7.1}{8.4}\selectfont NeurIPS-26} & 16.69 & 27.30 & 54.73 & 65.26 & 64.42 & 79.58 & 45.28 & 57.38 \\
\midrule[0.45pt]
\rowcolor{white} \cellcolor{white} & VTool-R1$^{*}$~\cite{wu2026vtoolr1} & {\fontsize{7.1}{8.4}\selectfont ICLR-26} & 20.49 & 33.88 & 52.59 & 67.04 & 64.16 & 78.32 & 45.75 & 59.75 \\
\rowcolor{tblstripe} \cellcolor{white}  & AdaReasoner$^{*}$~\cite{song2026adareasoner} & {\fontsize{7.1}{8.4}\selectfont ICLR-26} & 20.90 & 35.52 & 55.00 & 73.70 & 66.52 & 81.47 & 47.47 & 63.56 \\
\rowcolor{white} \cellcolor{white} \multirow{-3}{*}{\rotatebox{90}{Model}} & VisTA$^{*}$~\cite{huang2026vista} & {\fontsize{7.1}{8.4}\selectfont CVPR-26} & 17.76 & 26.78 & 49.91 & 59.63 & 59.62 & 72.03 & 42.43 & 52.81 \\
\midrule[0.45pt]
\rowcolor{tblstripe} \cellcolor{white} & {\fontsize{7.1}{8.4}\selectfont BetterTogether$^{*}$~\cite{soylu2024bettertogether}} & {\fontsize{7.1}{8.4}\selectfont EMNLP-24} & 21.45 & 31.69 & 57.69 & 70.00 & 64.42 & 77.97 & 47.85 & 59.89 \\
\rowcolor{white} \cellcolor{white} & {\fontsize{7.1}{8.4}\selectfont MetaTuner$^{*}$~\cite{bo2026metatuner}} & {\fontsize{7.1}{8.4}\selectfont ICLR-26} & 20.22 & 29.51 & 56.20 & 74.44 & 67.40 & 83.92 & 47.94 & 62.62 \\
\rowcolor{ours} \cellcolor{white} \multirow{-3}{*}{\rotatebox{90}{H+M}} & \textbf{\method} & {\fontsize{7.1}{8.4}\selectfont \textemdash} & \textbf{26.09} & \textbf{41.53} & \textbf{63.33} & \textbf{79.63} & \textbf{71.42} & \textbf{87.76} & \textbf{53.61} & \textbf{69.64} \\
\midrule[0.5pt]
\rowcolor{tblhead}\multicolumn{11}{l}{\rule{0pt}{10pt}\internvlicon \textbf{InternVL3.5-4B}}\\
\midrule[0.35pt]
\rowcolor{tblstripe} \cellcolor{white} & AWM~\cite{wang2024awm} & {\fontsize{7.1}{8.4}\selectfont ICML-25} & 17.97 & 31.37 & 52.78 & 66.37 & 56.26 & 75.16 & 42.34 & 57.63 \\
\rowcolor{white} \cellcolor{white} & DC~\cite{suzgun2025dc} & {\fontsize{7.1}{8.4}\selectfont EACL-26} & 17.19 & 33.55 & 42.54 & 57.02 & 55.79 & 74.95 & 38.51 & 55.17 \\
\rowcolor{white} \cellcolor{white} & XSkill~\cite{jiang2026xskill} & {\fontsize{7.1}{8.4}\selectfont ICML-26} & 18.34 & 35.53 & 42.87 & 56.12 & 57.63 & 78.11 & 39.61 & 56.59 \\
\rowcolor{tblstripe} \cellcolor{white} & GEPA~\cite{agrawal2025gepa} & {\fontsize{7.1}{8.4}\selectfont ICLR-26} & 18.32 & 35.64 & 48.55 & 60.58 & 59.11 & 77.47 & 41.99 & 57.90 \\
\rowcolor{white} \cellcolor{white} \multirow{-5}{*}{\rotatebox{90}{Harness}} & SkillOpt~\cite{yang2026skillopt} & {\fontsize{7.1}{8.4}\selectfont NeurIPS-26} & 21.23 & 45.21 & 48.66 & 61.92 & 58.26 & 78.32 & 42.72 & 61.82 \\
\midrule[0.45pt]
\rowcolor{white} \cellcolor{white} & VTool-R1$^{*}$~\cite{wu2026vtoolr1} & {\fontsize{7.1}{8.4}\selectfont ICLR-26} & 18.85 & 34.43 & 49.35 & 64.44 & 59.35 & 79.02 & 42.52 & 59.30 \\
\rowcolor{tblstripe} \cellcolor{white}  & AdaReasoner$^{*}$~\cite{song2026adareasoner} & {\fontsize{7.1}{8.4}\selectfont ICLR-26} & 17.08 & 37.16 & 53.52 & 69.63 & 60.93 & 81.82 & 43.84 & 62.87 \\
\rowcolor{white} \cellcolor{white} \multirow{-3}{*}{\rotatebox{90}{Model}} & VisTA$^{*}$~\cite{huang2026vista} & {\fontsize{7.1}{8.4}\selectfont CVPR-26} & 13.66 & 23.50 & 46.85 & 56.67 & 55.59 & 69.23 & 38.70 & 49.80 \\
\midrule[0.45pt]
\rowcolor{tblstripe} \cellcolor{white} & {\fontsize{7.1}{8.4}\selectfont BetterTogether$^{*}$~\cite{soylu2024bettertogether}} & {\fontsize{7.1}{8.4}\selectfont EMNLP-24} & 13.25 & 26.23 & 50.74 & 61.11 & 56.38 & 75.52 & 40.12 & 54.29 \\
\rowcolor{white} \cellcolor{white} & {\fontsize{7.1}{8.4}\selectfont MetaTuner$^{*}$~\cite{bo2026metatuner}} & {\fontsize{7.1}{8.4}\selectfont ICLR-26} & 12.57 & 24.59 & 53.61 & 67.41 & 62.33 & 81.12 & 42.83 & 57.71 \\
\rowcolor{ours} \cellcolor{white} \multirow{-3}{*}{\rotatebox{90}{H+M}} & \textbf{\method} & {\fontsize{7.1}{8.4}\selectfont \textemdash} & \textbf{21.86} & \textbf{45.90} & \textbf{58.98} & \textbf{75.56} & \textbf{66.08} & \textbf{87.41} & \textbf{48.97} & \textbf{69.62} \\
\midrule[0.5pt]
\rowcolor{tblhead}\multicolumn{11}{l}{\rule{0pt}{10pt}\internvlicon \textbf{InternVL3.5-8B}}\\
\midrule[0.35pt]
\rowcolor{tblstripe} \cellcolor{white} & AWM~\cite{wang2024awm} & {\fontsize{7.1}{8.4}\selectfont ICML-25} & 16.90 & 28.47 & 54.45 & 66.82 & 53.26 & 74.32 & 41.54 & 56.54 \\
\rowcolor{white} \cellcolor{white} & DC~\cite{suzgun2025dc} & {\fontsize{7.1}{8.4}\selectfont EACL-26} & 15.46 & 29.93 & 35.08 & 45.21 & 54.11 & 74.95 & 34.88 & 50.03 \\
\rowcolor{white} \cellcolor{white} & XSkill~\cite{jiang2026xskill} & {\fontsize{7.1}{8.4}\selectfont ICML-26} & 16.53 & 32.57 & 47.83 & 61.25 & 58.26 & 74.11 & 40.87 & 55.98 \\
\rowcolor{tblstripe} \cellcolor{white} & GEPA~\cite{agrawal2025gepa} & {\fontsize{7.1}{8.4}\selectfont ICLR-26} & 16.56 & 27.48 & 53.84 & 65.70 & 56.68 & 75.16 & 42.36 & 56.11 \\
\rowcolor{white} \cellcolor{white} \multirow{-5}{*}{\rotatebox{90}{Harness}} & SkillOpt~\cite{yang2026skillopt} & {\fontsize{7.1}{8.4}\selectfont NeurIPS-26} & 17.02 & 31.58 & 53.67 & 67.26 & 56.89 & 72.84 & 42.53 & 57.23 \\
\midrule[0.45pt]
\rowcolor{white} \cellcolor{white} & VTool-R1$^{*}$~\cite{wu2026vtoolr1} & {\fontsize{7.1}{8.4}\selectfont ICLR-26} & 13.52 & 27.32 & 54.63 & 68.52 & 58.65 & 77.27 & 42.27 & 57.70 \\
\rowcolor{tblstripe} \cellcolor{white}  & AdaReasoner$^{*}$~\cite{song2026adareasoner} & {\fontsize{7.1}{8.4}\selectfont ICLR-26} & 14.75 & 28.96 & 56.02 & 72.96 & 59.97 & 82.17 & 43.58 & 61.36 \\
\rowcolor{white} \cellcolor{white} \multirow{-3}{*}{\rotatebox{90}{Model}} & VisTA$^{*}$~\cite{huang2026vista} & {\fontsize{7.1}{8.4}\selectfont CVPR-26} & 12.02 & 20.77 & 51.39 & 62.59 & 53.32 & 68.88 & 38.91 & 50.75 \\
\midrule[0.45pt]
\rowcolor{tblstripe} \cellcolor{white} & {\fontsize{7.1}{8.4}\selectfont BetterTogether$^{*}$~\cite{soylu2024bettertogether}} & {\fontsize{7.1}{8.4}\selectfont EMNLP-24} & 9.84 & 19.13 & 55.37 & 69.26 & 56.21 & 73.78 & 40.47 & 54.05 \\
\rowcolor{white} \cellcolor{white} & {\fontsize{7.1}{8.4}\selectfont MetaTuner$^{*}$~\cite{bo2026metatuner}} & {\fontsize{7.1}{8.4}\selectfont ICLR-26} & 15.98 & 32.24 & 59.07 & 76.30 & 61.63 & 83.22 & 45.56 & 63.92 \\
\rowcolor{ours} \cellcolor{white} \multirow{-3}{*}{\rotatebox{90}{H+M}} & \textbf{\method} & {\fontsize{7.1}{8.4}\selectfont \textemdash} & \textbf{19.95} & \textbf{39.34} & \textbf{63.06} & \textbf{80.37} & \textbf{66.87} & \textbf{88.46} & \textbf{49.96} & \textbf{69.39} \\
\bottomrule[0.8pt]
\end{tabularx}
\endgroup
\end{table*}

\section{Experiments}
\label{sec:experiments}
\subsection{Experimental Setup}
\noindent\textbf{Datasets and metrics.}
\fullpar{The evaluation covers TIR-Bench~\cite{li2026tirbench}, MME-RealWorld-Lite~\cite{zhang2025mmerealworld}, and BLINK~\cite{fu2024blink}. Source tasks are divided into training, selection, and test splits, with image hashes and source identifiers keeping all derived views in their source partition. Average@4 measures mean correctness over four rollouts per task, and Pass@4 measures the fraction of tasks with at least one correct rollout. Both are reported in percent, with equal benchmark weights for macro averages. Paired 95\% confidence intervals use 10,000 bootstrap replicates over source tasks while retaining their four seeds together. Transformation analysis measures correct-map agreement, endpoint accuracy, and wrong-frame answers on applicable pairs. Appendix~\ref{sec:experimental_details} gives the splits and implementation settings.}

\noindent\textbf{Students and baselines.}
\fullpar[-1]{Students are Qwen3-VL-2B-Instruct, Qwen3-VL-8B-Instruct~\cite{bai2025qwen3vl}, InternVL3.5-4B, and InternVL3.5-8B~\cite{wang2025internvl35}. Qwen3-VL-8B is the reference student for ablations. Table~\ref{tab:main} groups baselines by the component they update. AWM~\cite{wang2024awm}, DC~\cite{suzgun2025dc}, XSkill~\cite{jiang2026xskill}, GEPA~\cite{agrawal2025gepa}, and SkillOpt~\cite{yang2026skillopt} refine external procedures. VTool-R1~\cite{wu2026vtoolr1}, AdaReasoner~\cite{song2026adareasoner}, and VisTA~\cite{huang2026vista} provide visual tool-learning comparisons, with VisTA retaining its separate selector. BetterTogether~\cite{soylu2024bettertogether} and MetaTuner~\cite{bo2026metatuner} optimize external prompts and model parameters. Their recipes are adapted to the declared backbones and tool interfaces, with auxiliary networks and calls included in resource accounting.}

\noindent\textbf{Implementation and controls.}
\fullpar{We use \texttt{gpt-5.6-sol} as the teacher. Model-RSI uses AdamW~\cite{loshchilov2019adamw} to train LoRA adapters~\cite{hu2022lora} in attention query, key, value, and output projections. The pretrained weights and selected harness remain fixed during training. Comparisons share initial students, training sources, image settings, tool implementations, and inference limits, with method-specific budget differences recorded. Matched-data controls use a common verified pool and fixed harness. Appendix~\ref{sec:augmentation_controls} reports equal source-step and equal training-FLOP controls, including paired branches and candidate scoring. Appendix~\ref{sec:internalization} reports offline costs for generation, selection, and optimization across accepted and rejected updates.}

\subsection{Main Results}
\fullpar{\noindent\textbf{Comparison across backbones.} Table~\ref{tab:main} shows that \method achieves the highest reported Average@4 and Pass@4 on every backbone--benchmark pair, spanning both model families and all tested sizes. Macro Average@4 improves by 4.40--5.67 percentage points over the strongest reported baseline for each backbone.}

\fullpar{\noindent\textbf{Comparison across optimization strategies.} Among harness baselines, SkillOpt approaches \method on TIR-Bench with InternVL3.5-4B, but the gap is wider on MME-RealWorld-Lite and BLINK. \method also outperforms model and joint optimization baselines, including BetterTogether and MetaTuner, which already update prompts and model parameters. Our framework connects the two loops through visual experience: harness executions supply verified views, actions, and relations for model learning, while updated models guide subsequent harness refinement. The results are consistent with this design; the ablations below examine individual component contributions.}

\subsection{Ablation Studies}
\begin{table*}[!t]
\centering
\caption{Core-component ablation with Qwen3-VL-8B. Results report Average@4 and Pass@4 in percent ($\uparrow$) on the same benchmarks as Table~\ref{tab:main}. Macro Avg. weights the three benchmarks equally. The full configuration is highlighted in red.}
\label{tab:ablation}
\begingroup\tablelook
\fontsize{8}{9.5}\selectfont
\setlength{\tabcolsep}{2pt}
\renewcommand{\arraystretch}{1.2}
\begin{tabularx}{\textwidth}{>{\centering\arraybackslash}p{0.04\textwidth}|*{5}{>{\centering\arraybackslash}p{0.047\textwidth}}||YY|YY|YY|YY}
\toprule[0.8pt]
\rowcolor{tblhead}&\multicolumn{5}{c||}{\textbf{Components}}&\multicolumn{2}{c|}{\textbf{TIR-Bench}}&\multicolumn{2}{c|}{\textbf{MME-RealWorld-Lite}}&\multicolumn{2}{c|}{\textbf{BLINK}}&\multicolumn{2}{c}{\textbf{Macro Avg.}}\\
\rowcolor{tblhead}\textbf{Index}&\textbf{H}&\textbf{M}&\textbf{Curr.}&\textbf{Rel.}&\textbf{Comp.}&Avg.@4&Pass@4&Avg.@4&Pass@4&Avg.@4&Pass@4&Avg.@4&Pass@4\\
\midrule[0.45pt]
\rowcolor{white}1&--&\yes&\yes&\yes&\yes&22.81&36.07&57.59&72.22&66.70&82.52&49.03&63.60\\
\rowcolor{tblstripe}2&\yes&--&--&--&--&21.58&33.33&55.28&67.78&64.51&80.07&47.12&60.39\\
\rowcolor{white}3&\yes&\yes&--&\yes&\yes&24.59&38.25&59.07&74.44&69.06&84.97&50.91&65.89\\
\rowcolor{tblstripe}4&\yes&\yes&\yes&--&\yes&23.77&37.16&60.46&75.56&68.09&83.92&50.78&65.54\\
\rowcolor{white}5&\yes&\yes&\yes&\yes&--&25.82&40.98&63.06&79.26&71.07&87.41&53.31&69.22\\
\rowcolor{ours}\textbf{6}&\yes&\yes&\yes&\yes&\yes&\textbf{26.09}&\textbf{41.53}&\textbf{63.33}&\textbf{79.63}&\textbf{71.42}&\textbf{87.76}&\textbf{53.61}&\textbf{69.64}\\
\bottomrule[0.8pt]
\end{tabularx}
\par\smallskip
\parbox{\textwidth}{\footnotesize H: harness refinement; M: model updates; Curr.: curriculum ordering; Rel.: visual relation supervision; Comp.: harness compaction. Checks mark active components; dashes mark disabled or inapplicable components. Disabling model updates also disables curriculum, relation losses, and post-model compaction.}
\endgroup
\end{table*}

\fullpar{\noindent\textbf{Harness and model updates.} Table~\ref{tab:ablation} compares the full method with five component-removal variants. Removing model updates causes the largest decline, indicating that harness search alone cannot recover the benefit of learning from visual experience. Removing harness refinement also hurts every benchmark, supporting continued adaptation of external procedures. Without model updates, curriculum, relation losses, and post-model compaction are also disabled.}

\fullpar{\noindent\textbf{Curriculum and relation supervision.} Randomizing the same verified examples lowers performance despite matched exposure counts and optimizer steps, supporting progressive admission as a useful way to organize supervision. Removing relation losses also hurts performance while retaining answer and action targets. This comparison supports learning dependencies between views in addition to supervising each view individually. The two ablations exhibit similar macro-level declines, consistent with contributions from both training order and relational structure.}

\fullpar{\noindent\textbf{Harness compaction.} Removing compaction has the smallest effect, with performance remaining close to the full configuration. This pattern is consistent with its role in consolidating external guidance after model learning.}

\begin{figure*}[!t]
\centering
\includegraphics[width=\textwidth]{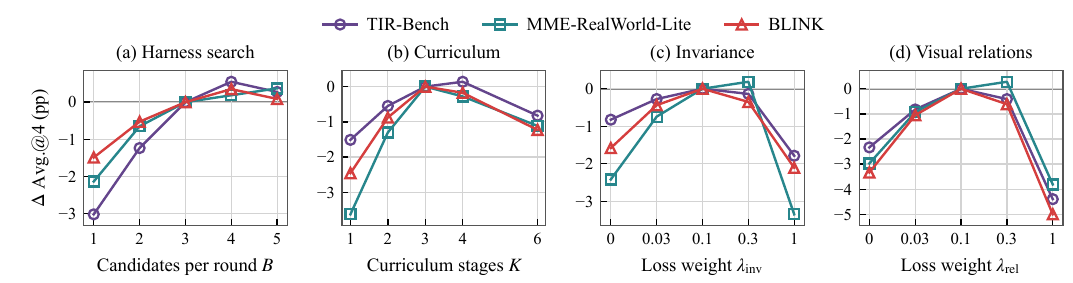}
\caption{Hyperparameter sensitivity of Qwen3-VL-8B on TIR-Bench, MME-RealWorld-Lite, and BLINK. Panels vary (a) candidates per round $B$, (b) curriculum stages $K$, (c) invariance weight $\lambda_{\mathrm{inv}}$, and (d) visual relation weight $\lambda_{\mathrm{rel}}$. Vertical axes show validation Average@4 changes (percentage points) relative to $B=3$, $K=3$, and $\lambda_{\mathrm{inv}}=\lambda_{\mathrm{rel}}=0.1$. Discrete loss weights use equally spaced positions.}
\label{fig:hyperparameters}
\end{figure*}

\Needspace{4\baselineskip}
\subsection{Further Analysis}
\fullpar{\noindent\textbf{Hyperparameter sensitivity.} Figure~\ref{fig:hyperparameters} suggests trade-offs between exploration, curriculum granularity, and supervision balance. Increasing $B$ gives selection more alternative harness edits, but gains from further evaluations diminish as selection can reject additional candidates. The preference for $K=3$--4 is consistent with gradually introducing harder views: too few stages weaken this progression, whereas too many may delay access to challenging examples. Earlier views remain eligible after admission, so later stages broaden supervision without discarding the easier examples. The schedule changes access to fixed targets rather than relabeling them. Relation weights control how strongly paired predictions constrain one another beyond answer and action supervision. Small weights provide weak coupling; large weights may overwhelm answer supervision. These coefficients adjust the balance of training signals while the verified targets and relation mappings remain fixed. The favorable performance near $\lambda_{\mathrm{inv}}=\lambda_{\mathrm{rel}}=0.1$ supports a moderate contribution from relational constraints.}

\Needspace{3\baselineskip}
\noindent\textbf{Transformation interpretation.}
\fullpar[-1]{Figure~\ref{fig:motivation} illustrates how a recorded crop links local and global answers. The same target occupies 40\% of the crop and 8\% of the full image, with the crop footprint determining the conversion. Qualitative comparisons align the source image, tool arguments, returned observation, requested frame, next action, and final answer to reveal where interpretation changes along an execution.}

\section{Discussion}
\fullpar{Visual tools expose relationships between observations as well as new task information. Learning these relationships provides a common principle for interpreting local measurements, converting coordinates, and composing visual operations. \method connects this principle to two complementary forms of adaptation: reusable execution guidance and learned model behavior. Extending this approach to richer tool libraries and longer workflows requires reliable transformation semantics for supervision and update selection. The formulation assumes that a recorded operation supports a verifiable answer mapping. Occlusion, incomplete tool returns, or ambiguous reference frames can invalidate that assumption; such examples retain only independently verified supervision. Moreover, repeated acceptance tests can adapt to a finite selection set. Improvements on that set therefore require separate held-out evaluation before being interpreted as broader generalization.}

\fullpar{Relation supervision depends on the question as well as the image operation. A crop can preserve a recognition target while changing a count or an area denominator. The same operation therefore need not define the same relation across tasks. Explicitly recording the requested quantity and reference frame makes this distinction testable and helps determine whether paired supervision is justified. It also separates failures to recognize visual evidence from failures to convert an otherwise correct measurement into the requested answer frame.}

\section{Conclusion}
\fullpar[-1]{We introduce \method for learning structured visual relations from multimodal tool interactions. The method combines verified answer and action targets with answer-preserving consistency, transformation mappings, and wrong-map contrast. A progressive curriculum connects these signals to alternating harness refinement and parameter learning, with each update evaluated under a fixed counterpart. This framework turns recorded visual operations into reusable supervision for multimodal agents, linking external guidance to learned visual capabilities.}

\section*{LLM Usage Statement}
\fullpar{OpenAI ChatGPT assisted with author-produced text, related work and citations, code, experimental design and analysis, figures and visual material, data-related material, research ideas, and reviewer-style manuscript feedback. OpenAI Codex assisted with LaTeX template migration, manuscript language polishing, paragraph and caption layout, PDF compilation checks, and this disclosure. These tools are distinct from the models evaluated in our method. The authors take full responsibility for the accuracy, originality, and integrity of all submitted content.}

{\small
\let\originalthebibliography\thebibliography
\renewcommand{\thebibliography}[1]{%
  \originalthebibliography{#1}%
  \setlength{\parfillskip}{0pt plus 0.18\linewidth}}
\bibliographystyle{teachrsi_refs}
\bibliography{references}
}

\clearpage
\appendix
\counterwithin{table}{section}
\counterwithin{figure}{section}
\counterwithin{equation}{section}
\renewcommand{\thcell}[1]{\textbf{\shortstack[c]{\strut #1\strut}}}
\captionsetup{singlelinecheck=false,hypcap=false}
\pdfbookmark[0]{Appendix}{appendix}
\section*{Appendix}
\noindent\textbf{Scope.}
The appendix reports experimental settings, repeated runs, relation ablations, transformation transfer, recursive schedules, harness diagnostics, and supervision audits. Scores are percentages unless stated otherwise; pale-red rows identify the complete configuration.

\section{Experimental Details}
\label{sec:experimental_details}
\noindent\textbf{Tasks and students.}
We evaluate TIR-Bench, MME-RealWorld-Lite, and BLINK with Qwen3-VL-2B-Instruct, Qwen3-VL-8B-Instruct, InternVL3.5-4B, and InternVL3.5-8B. Qwen3-VL-8B is the reference ablation student, and the teacher is \texttt{gpt-5.6-sol}. Tables abbreviate the three benchmarks as TIR, MME, and BLINK.

\noindent\textbf{Source partitions.}
Table~\ref{tab:common_eval_split} gives the common source-task split: approximately 45\% training, 15\% selection, and 40\% testing. The training partition supports trajectory collection and verified supervision; the selection partition supports harness-candidate and checkpoint decisions. Questions and derived views linked to the same image or source are assigned together, with task-type stratification where feasible. All methods and backbones share the same held-out tasks and four rollout seeds per source task.

\noindent\textbf{Test strata.}
The held-out set has two disjoint strata. Test A contains tasks with a verifiable reference-changing mapping, the capability targeted by relation supervision; Test B contains the remaining tasks. Benchmark scores weight the two strata equally, and each source task has equal weight within its stratum.

\noindent\textbf{Reported metrics.}
The mean correctness and task-success fraction of main-paper Section~4.1 are computed within each stratum. Benchmark-level Average@4 and Pass@4 in main-paper Tables~1 and~2 are the equal-weight mean of the Test A and Test B scores; Macro columns average the three benchmarks equally. Section~\ref{sec:multirun} gives the formulas.

\par\medskip\noindent\begin{minipage}{\linewidth}
\centering
\captionof{table}{Common source-task split for all methods and backbones. Test A and Test B are disjoint and weighted equally in reported accuracy.}
\label{tab:common_eval_split}
\begingroup\tablelook\fontsize{8}{9.5}\selectfont
\begin{tabularx}{\linewidth}{L{0.16\linewidth}|YYYYY}\toprule[0.65pt]
\rowcolor{tblhead}\textbf{Data}&\textbf{Train}&\textbf{Select}&\textbf{Test A}&\textbf{Test B}&\thcell{Test\\total}\\\midrule[0.4pt]
\rowcolor{white}TIR&546&182&183&304&\textbf{487}\\
\rowcolor{tblstripe}MME&808&269&270&449&\textbf{719}\\
\rowcolor{white}BLINK&855&285&286&475&\textbf{761}\\
\bottomrule[0.65pt]\end{tabularx}\endgroup
\end{minipage}\par\medskip

\noindent\textbf{Selection screen.}
Harness candidates are evaluated on a fixed, stratified 81-task screen drawn from the selection partition (20 TIR, 30 MME, 31 BLINK), with four rollouts per task. The same screen is used for candidate and incumbent evaluations; Table~\ref{tab:selection_ledger_review} reports both event types.

\noindent\textbf{Model adaptation.}
Model-RSI trains LoRA adapters on the language-decoder attention query, key, value, and output projections with AdamW. The vision encoder, projector, embeddings, and remaining pretrained parameters are frozen, and base weights are not quantized. Table~\ref{tab:training_settings} lists training and search settings. LoRA weights carry forward between rounds; optimizer and scheduler states restart for each candidate update. Baselines retain their method-specific algorithms, and shared settings apply to the corresponding adaptations and matched-data controls.

\par\medskip\noindent\begin{minipage}{\linewidth}
\centering
\captionof{table}{Evaluation, training, and search settings. Steps and batch counts refer to source tasks.}
\label{tab:training_settings}
\begingroup\tablelook\fontsize{8}{9.5}\selectfont
\begin{tabularx}{\linewidth}{L{0.40\linewidth}|>{\raggedright\arraybackslash}X}
\toprule[0.65pt]
\rowcolor{tblhead}\textbf{Setting}&\textbf{Value}\\
\midrule[0.4pt]
\rowcolor{tblstripe}Evaluation rollouts&4 per source task\\
\rowcolor{white}Cluster GPU&NVIDIA A100-SXM4; 80\,GB per GPU\\
\rowcolor{tblstripe}Distributed inference worker&2 GPUs; tensor parallel size 2\\
\rowcolor{white}Core runtime&PyTorch~2.9.0; vLLM~0.11.2\\
\rowcolor{tblstripe}GPU runtime libraries&CUDA runtime~12.8.90; cuDNN~9.10.2.21; NCCL~2.27.5\\
\rowcolor{white}Trainable projections&Attention query/key/value/output\\
\rowcolor{tblstripe}LoRA rank / alpha / dropout&32 / 64 / 0.05\\
\rowcolor{white}Optimizer&AdamW\\
\rowcolor{tblstripe}Learning rate / decay&$2\!\times\!10^{-5}$ / 0.01\\
\rowcolor{white}Adam betas / epsilon&$(0.9,0.999)$ / $10^{-8}$\\
\rowcolor{tblstripe}Schedule / warm-up&Cosine / 3\% of update steps\\
\rowcolor{white}Precision / gradient clip&BF16 / global norm 1.0\\
\rowcolor{tblstripe}Global source batch&16; microbatch 1 per device\\
\rowcolor{white}Round / search schedule&3 rounds; 6 search steps/round\\
\rowcolor{tblstripe}Candidate limit $B$&3 per search step\\
\rowcolor{white}Model-update schedule&600 steps/round; 200/stage\\
\rowcolor{tblstripe}Curriculum stages $K$&3\\
\rowcolor{white}Admission thresholds&Ordinal difficulty 1, 2, 3\\
\rowcolor{tblstripe}$\lambda_{\rm inv}$ / $\lambda_{\rm rel}$&0.1 / 0.1\\
\rowcolor{white}$\omega$ / $\gamma$ / $\beta$&1.0 / 1.0 / 0.1\\
\bottomrule[0.65pt]
\end{tabularx}\endgroup
\end{minipage}\par\medskip

\noindent\textbf{Compute environment.}
Experiments run as independent jobs on a cluster of NVIDIA A100-SXM4 GPUs (80\,GB each). Distributed student inference uses two GPUs per worker with tensor parallel size 2. The runtime uses PyTorch~2.9.0 and vLLM~0.11.2, with CUDA runtime~12.8.90, cuDNN~9.10.2.21, and NCCL~2.27.5. Total concurrent GPU use depends on the active jobs; training and per-worker inference allocations are accounted for separately in measured GPU-hours.

\noindent\textbf{Exposure and compute.}
The reference schedule has 18 harness-search steps, at most 54 proposal slots, and 1,800 optimizer updates per complete run, before any model-update rejection. Invalid or duplicate proposals consume slots, and rejected updates retain their actual training cost. With $g$ devices and one source per device per microbatch, accumulation is $16/g$ when integral; otherwise a compatible microbatch and accumulation pair maintains the global source batch. Relation branches increase work within a source without changing that batch. Equal-source-step controls preserve 1,800 updates and source exposure; equal-FLOP controls adjust update counts to a common measured compute ceiling. One-pass and different-round schedules redistribute the total search and update budgets across rounds. Realized teacher calls, training compute, and selection executions are reported separately.

\noindent\textbf{Views, curriculum, and source sampling.}
Each source retains its original view and up to three verified augmented views. Cropping, resizing, contrast adjustment, and rotation are admitted only when the question has supported evidence and valid answer semantics. Difficulty 1 denotes visible evidence and direct reasoning; difficulty 2 introduces distractors or moderate scale or frame conversion; difficulty 3 includes small targets, clutter, or multi-step reasoning while retaining sufficient evidence. The teacher assigns these labels before optimization. Three cumulative stages admit levels $\le1$, $\le2$, and $\le3$, retaining earlier views. Active source tasks are sampled uniformly, followed by at most two admitted answer views, one verified action target, one invariant pair, and one reference-changing pair per source and update when available. Missing groups contribute zero according to the objective. Pair endpoints count toward realized compute.

\noindent\textbf{Answer domains and mappings.}
The objective uses canonical candidate strings, column-stochastic transport matrices, and correct-map fitting with wrong-map contrast. Scalar area fractions and normalized single-axis coordinates use a $0.01$ grid in $[0,1]$. Candidates are scored by mean answer-token log probability and normalized by softmax at temperature 1. Exact candidate matches transfer probability mass directly; intermediate mapped values use linear interpolation between adjacent valid bins. Out-of-domain relations are rejected rather than clipped. Coordinates retain explicit units and reference dimensions; multidimensional answers and counts use task-specific candidate domains. Each reference-changing pair uses up to three valid distinct wrong maps, such as omitted offset, omitted scale, or local-as-global identity. Alternatives agreeing with the target within the answer tolerance are excluded. The numerical tie tolerance is $10^{-6}$; geometric validity and answer tolerances are task-specific.

\noindent\textbf{Inference, prompts, and resource accounting.}
Table~\ref{tab:execution_settings} gives decoding and execution settings. Each backbone uses its supported image processor and native aspect-ratio handling. Prompt versions distinguish harness refinement, verification, difficulty labeling, and compaction. Online costs include failures and average the four rollout costs within each task rather than summing them. Section~\ref{sec:accounting_review} defines logical unique input, cumulative full-prompt input, and the distinction from physical cache work. Wall time, peak memory, and token usage are measured separately from static harness length.

\par\medskip\noindent\begin{minipage}{\linewidth}
\centering
\captionof{table}{Decoding and execution settings. Output and tool limits apply to each rollout unless stated otherwise.}
\label{tab:execution_settings}
\begingroup\tablelook\fontsize{8}{9.5}\selectfont
\begin{tabularx}{\linewidth}{L{0.40\linewidth}|X}
\toprule[0.65pt]
\rowcolor{tblhead}\textbf{Setting}&\textbf{Value}\\
\midrule[0.4pt]
\rowcolor{white}Teacher proposal sampling&Temperature 0.7, top-$p$ 0.95\\
\rowcolor{tblstripe}Teacher verification sampling&Temperature 0; independent audit retained\\
\rowcolor{white}Teacher output limit&8,192 tokens per response\\
\rowcolor{tblstripe}Student rollout sampling&Temperature 0.7, top-$p$ 0.95\\
\rowcolor{white}Student output budget&4,096 generated tokens/rollout, accumulated across assistant turns\\
\rowcolor{tblstripe}Tool limit / timeout&8 calls/rollout; 30 seconds/call\\
\rowcolor{white}Rollout wall-time limit&180 seconds, including model and tools\\
\rowcolor{tblstripe}Selection checkpoint&End of each 600-step model update; accept only positive matched gain\\
\rowcolor{white}Harness acceptance&Strict positive matched score gain; deterministic first-in-order tie rule\\
\rowcolor{tblstripe}Compaction&One proposal after each accepted model update; shorter, non-degrading guidance only\\
\rowcolor{white}Scoring&Task-native parser and numerical answer tolerances\\
\bottomrule[0.65pt]
\end{tabularx}\endgroup
\end{minipage}\par\medskip

\section{Main-Experiment Multi-Run Results}
\label{sec:multirun}
\noindent\textbf{Reference identity and repeat scope.}
The main-paper TeachRSI$^2$ point estimate for each student is denoted R0; Qwen3-VL-8B has R0 macro Average@4 of 53.61. Tables~\ref{tab:main_multirun} and~\ref{tab:perrun_results} summarize R0 together with R1--R3. The main table reports the reference evaluation, and the supplement reports four-run summaries. R1--R3 repeat harness search, generation, training, and acceptance under the same configuration. MetaTuner uses four reported runs for matched repeat counts.

\noindent\textbf{Fixed-harness controls.}
The fixed-harness ablations use the shared reference $H_{\mathrm{abl}}$ defined in Section~\ref{sec:relation_ablation}, with a common verified snapshot, source partitions, and target mappings. Their variability is distinct from end-to-end search and data generation. Tables~\ref{tab:main_multirun} and~\ref{tab:perrun_results} report run dispersion; Tables~\ref{tab:primary_ci} and~\ref{tab:secondary_ci} report task-sampling intervals for the listed contrasts.

\par\medskip\noindent\begin{minipage}{\linewidth}
\centering
\captionof{table}{Random-seed schedule for R0--R3. Separate namespaces control training, search sampling, and generation bookkeeping. Split and evaluation seeds are fixed across runs.}
\label{tab:seed_schedule}
\begingroup\tablelook\fontsize{8}{9.5}\selectfont
\begin{tabularx}{\linewidth}{L{0.13\linewidth}|YYY}
\toprule[0.65pt]
\rowcolor{tblhead}\textbf{Run}&\textbf{Train}&\textbf{Search}&\textbf{Generate}\\
\midrule[0.4pt]
\rowcolor{white}R0&17&1017&2017\\
\rowcolor{tblstripe}R1&11&1011&2011\\
\rowcolor{white}R2&22&1022&2022\\
\rowcolor{tblstripe}R3&33&1033&2033\\
\bottomrule[0.65pt]
\end{tabularx}
\par\smallskip\raggedright\footnotesize Split seed: 2026. Evaluation rollout seeds: 42, 43, 44, 45 for every run. R1--R3 denote the additional repeats; R0 denotes the main-paper reference evaluation.\par
\endgroup\end{minipage}\par\medskip

\noindent\textbf{Sources of variability.}
Separate seed namespaces control training, search sampling, generation bookkeeping, and evaluation. Training and search randomness, together with hosted-teacher sampling, contribute to the observed variation across runs.

\noindent\textbf{Within-run metrics.}
For run $r$, benchmark $b$, and stratum $s\in\{A,B\}$ containing $N_{b,s}$ source tasks, let $z_{r,b,s,i,k}\in\{0,1\}$ indicate correctness on rollout $k$. Empty or unparseable model answers count as incorrect and remain in the denominator. First compute the task-equal metrics within each stratum:
\begin{equation}
\mathrm{Avg}_{r,b,s}@4=\frac{100}{4N_{b,s}}\sum_{i=1}^{N_{b,s}}\sum_{k=1}^{4}z_{r,b,s,i,k},
\end{equation}
\begin{equation}
\mathrm{Pass}_{r,b,s}@4=\frac{100}{N_{b,s}}\sum_{i=1}^{N_{b,s}}
\mathbf 1\!\left[\sum_{k=1}^{4}z_{r,b,s,i,k}>0\right].
\end{equation}
For either metric $m$, the reported benchmark and macro scores are
\begin{equation}
\begin{aligned}
m_{r,b}&=\tfrac12(m_{r,b,A}+m_{r,b,B}),\\
m_{r,\mathrm{macro}}&=\tfrac13\sum_{b=1}^{3}m_{r,b}.
\end{aligned}
\end{equation}

\noindent\textbf{Common aggregation definition.}
Each run's macro score is the equal-weight mean of its three benchmark scores. Repeated-run results use the same run-first definition for every method: compute Average@4 and Pass@4 within each run, compute its benchmark macro score, and then aggregate the run-level scores. For a method reported over $R\ge2$ recorded runs, the mean and sample standard deviation of either metric $m_r$ are
\begin{equation}
\bar m=\frac{1}{R}\sum_{r=1}^{R}m_r,\qquad
s_m=\sqrt{\frac{\sum_{r=1}^{R}(m_r-\bar m)^2}{R-1}},
\end{equation}
Here $R$ is the actual number of runs summarized by the reported result; Tables~\ref{tab:main_multirun} and~\ref{tab:perrun_results} use $R=4$. Pass@4 is computed within each run before aggregation, retaining its four-rollout definition. Single-run reference results retain their within-run scores and are labeled separately from repeated-run means and standard deviations.

\noindent\textbf{Reported uncertainty.}
Tables~\ref{tab:main_multirun} and~\ref{tab:perrun_results} report four-run means and sample SD. Tables~\ref{tab:primary_ci} and~\ref{tab:secondary_ci} report paired differences and 95\% task-bootstrap intervals. Section~\ref{sec:paired_ci} defines their statistical scope and interpretation.

\par\medskip\noindent\begin{minipage}{\linewidth}
\centering
\captionof{table}{Four-run main-experiment summary: R0 from main-paper Table 1 and R1--R3 from the repeat panel. Values are means $\pm$ sample SD, separately for each method. Paired method-difference intervals are in Table~\ref{tab:primary_ci}.}
\label{tab:main_multirun}
\begingroup\tablelook\fontsize{8}{9.5}\selectfont
\setlength{\tabcolsep}{2pt}
\begin{tabularx}{\linewidth}{L{0.26\linewidth}|L{0.19\linewidth}|YY}
\toprule[0.65pt]
\rowcolor{tblhead}\textbf{Student}&\textbf{Method}&\thcell{Macro\\Avg.@4}&\thcell{Macro\\Pass@4}\\
\rowcolor{white}Qwen3-VL-2B&MetaTuner&\uncert{44.46}{0.35}&\uncert{65.92}{0.21}\\
\rowcolor{ours}Qwen3-VL-2B&\method&\uncert{49.85}{0.12}&\uncert{71.70}{0.50}\\
\rowcolor{white}Qwen3-VL-8B&MetaTuner&\uncert{48.00}{0.18}&\uncert{62.90}{0.32}\\
\rowcolor{ours}Qwen3-VL-8B&\method&\uncert{53.75}{0.20}&\uncert{69.53}{0.72}\\
\rowcolor{white}InternVL3.5-4B&MetaTuner&\uncert{42.73}{0.26}&\uncert{57.60}{0.45}\\
\rowcolor{ours}InternVL3.5-4B&\method&\uncert{49.04}{0.29}&\uncert{69.64}{0.41}\\
\rowcolor{white}InternVL3.5-8B&MetaTuner&\uncert{45.57}{0.15}&\uncert{63.76}{0.51}\\
\rowcolor{ours}InternVL3.5-8B&\method&\uncert{49.85}{0.22}&\uncert{69.38}{0.84}\\

\bottomrule[0.65pt]
\end{tabularx}\endgroup\end{minipage}\par\medskip
\par\medskip

\par\medskip\noindent\begin{minipage}{\linewidth}
\centering
\captionof{table}{Four-run TeachRSI$^2$ results for all students. Cells show Average@4 / Pass@4 (\%); final cells give each metric\textquotesingle s mean $\pm$ sample SD. R0 is the main-table reference; R1--R3 are the additional repeats. }
\label{tab:perrun_results}
\begingroup\tablelook\fontsize{8}{9.5}\selectfont
\setlength{\tabcolsep}{2pt}
\begin{tabularx}{\linewidth}{L{0.11\linewidth}|YYYY|>{\centering\arraybackslash}p{0.24\linewidth}}
\toprule[0.65pt]
\midrule[0.35pt]\rowcolor{tblhead}\multicolumn{6}{l}{\textbf{Qwen3-VL-2B with TeachRSI$^2$}}\\
\rowcolor{tblhead}\textbf{Data}&\textbf{R0}&\textbf{R1}&\textbf{R2}&\textbf{R3}&\textbf{Mean $\pm$ SD}\\
\rowcolor{white}TIR&\scorepair{27.73}{56.83}&\scorepair{28.14}{56.83}&\scorepair{28.55}{57.38}&\scorepair{27.32}{57.38}&\scorepair{\uncert{27.94}{0.53}}{\uncert{57.10}{0.32}}\\
\rowcolor{white}MME&\scorepair{58.15}{74.07}&\scorepair{58.89}{73.33}&\scorepair{57.69}{73.70}&\scorepair{58.80}{75.19}&\scorepair{\uncert{58.38}{0.57}}{\uncert{74.07}{0.80}}\\
\rowcolor{white}BLINK&\scorepair{63.11}{84.27}&\scorepair{62.76}{83.57}&\scorepair{63.37}{83.22}&\scorepair{63.64}{84.62}&\scorepair{\uncert{63.22}{0.37}}{\uncert{83.92}{0.64}}\\
\rowcolor{ours}Macro&\scorepair{49.66}{71.72}&\scorepair{49.93}{71.24}&\scorepair{49.87}{71.43}&\scorepair{49.92}{72.39}&\scorepair{\uncert{49.85}{0.12}}{\uncert{71.70}{0.50}}\\
\midrule[0.35pt]\rowcolor{tblhead}\multicolumn{6}{l}{\textbf{Qwen3-VL-8B with TeachRSI$^2$}}\\
\rowcolor{tblhead}\textbf{Data}&\textbf{R0}&\textbf{R1}&\textbf{R2}&\textbf{R3}&\textbf{Mean $\pm$ SD}\\
\rowcolor{white}TIR&\scorepair{26.09}{41.53}&\scorepair{26.50}{40.44}&\scorepair{26.91}{40.44}&\scorepair{26.09}{42.08}&\scorepair{\uncert{26.40}{0.39}}{\uncert{41.12}{0.82}}\\
\rowcolor{white}MME&\scorepair{63.33}{79.63}&\scorepair{63.80}{79.26}&\scorepair{62.96}{78.89}&\scorepair{64.07}{80.74}&\scorepair{\uncert{63.54}{0.49}}{\uncert{79.63}{0.80}}\\
\rowcolor{white}BLINK&\scorepair{71.42}{87.76}&\scorepair{70.89}{88.46}&\scorepair{70.98}{86.71}&\scorepair{71.94}{88.46}&\scorepair{\uncert{71.31}{0.48}}{\uncert{87.85}{0.83}}\\
\rowcolor{ours}Macro&\scorepair{53.61}{69.64}&\scorepair{53.73}{69.39}&\scorepair{53.62}{68.68}&\scorepair{54.04}{70.43}&\scorepair{\uncert{53.75}{0.20}}{\uncert{69.53}{0.72}}\\
\midrule[0.35pt]\rowcolor{tblhead}\multicolumn{6}{l}{\textbf{InternVL3.5-4B with TeachRSI$^2$}}\\
\rowcolor{tblhead}\textbf{Data}&\textbf{R0}&\textbf{R1}&\textbf{R2}&\textbf{R3}&\textbf{Mean $\pm$ SD}\\
\rowcolor{white}TIR&\scorepair{21.86}{45.90}&\scorepair{21.99}{45.90}&\scorepair{21.45}{45.90}&\scorepair{22.27}{46.45}&\scorepair{\uncert{21.89}{0.34}}{\uncert{46.04}{0.27}}\\
\rowcolor{white}MME&\scorepair{58.98}{75.56}&\scorepair{59.44}{75.56}&\scorepair{58.43}{75.19}&\scorepair{59.44}{75.93}&\scorepair{\uncert{59.07}{0.48}}{\uncert{75.56}{0.30}}\\
\rowcolor{white}BLINK&\scorepair{66.08}{87.41}&\scorepair{66.35}{87.41}&\scorepair{66.08}{86.36}&\scorepair{66.08}{88.11}&\scorepair{\uncert{66.15}{0.13}}{\uncert{87.33}{0.72}}\\
\rowcolor{ours}Macro&\scorepair{48.97}{69.62}&\scorepair{49.26}{69.62}&\scorepair{48.65}{69.15}&\scorepair{49.27}{70.16}&\scorepair{\uncert{49.04}{0.29}}{\uncert{69.64}{0.41}}\\
\midrule[0.35pt]\rowcolor{tblhead}\multicolumn{6}{l}{\textbf{InternVL3.5-8B with TeachRSI$^2$}}\\
\rowcolor{tblhead}\textbf{Data}&\textbf{R0}&\textbf{R1}&\textbf{R2}&\textbf{R3}&\textbf{Mean $\pm$ SD}\\
\rowcolor{white}TIR&\scorepair{19.95}{39.34}&\scorepair{19.40}{40.44}&\scorepair{19.54}{38.25}&\scorepair{19.54}{39.89}&\scorepair{\uncert{19.60}{0.24}}{\uncert{39.48}{0.93}}\\
\rowcolor{white}MME&\scorepair{63.06}{80.37}&\scorepair{63.80}{81.11}&\scorepair{63.80}{79.26}&\scorepair{62.50}{80.37}&\scorepair{\uncert{63.29}{0.63}}{\uncert{80.28}{0.76}}\\
\rowcolor{white}BLINK&\scorepair{66.87}{88.46}&\scorepair{66.35}{89.51}&\scorepair{66.87}{87.41}&\scorepair{66.61}{88.11}&\scorepair{\uncert{66.67}{0.25}}{\uncert{88.37}{0.87}}\\
\rowcolor{ours}Macro&\scorepair{49.96}{69.39}&\scorepair{49.85}{70.35}&\scorepair{50.07}{68.31}&\scorepair{49.55}{69.46}&\scorepair{\uncert{49.85}{0.22}}{\uncert{69.38}{0.84}}\\

\bottomrule[0.65pt]
\end{tabularx}\endgroup\end{minipage}\par\medskip\par\medskip

\noindent\textbf{Four-run comparison.}
The four-run macro Average@4 advantages over MetaTuner are 5.39, 5.75, 6.31, and 4.29 percentage points for Qwen3-VL-2B, Qwen3-VL-8B, InternVL3.5-4B, and InternVL3.5-8B. Pass@4 differences are 5.78, 6.64, 12.04, and 5.62 points. All eight paired intervals in Table~\ref{tab:primary_ci} are above zero, supporting the improvement across both model families.

\section{Disentangling Visual Relation Supervision}
\label{sec:relation_ablation}
\noindent\textbf{Budget matching.}
For the fixed-harness studies in Tables~\ref{tab:relation_results}, \ref{tab:generalization}, \ref{tab:augmentation_results}, and~\ref{tab:compute_controls}, $H_{\mathrm{abl}}=H_1^+$ is the round-1 accepted, pre-compaction harness (2,917 tokens; Table~\ref{tab:round_history_review}), frozen immediately after the first Harness-RSI stage and before the first Model-RSI update. Compared variants share the initial student checkpoint $\theta_0$ and the verified training snapshot $\mathcal Z$ collected under $(\theta_0,H_1^+)$. The harness remains unchanged throughout training and evaluation; transformation-generalization methods share the same evaluation population within each condition. Equal-source-step comparisons match source exposure and optimizer updates, while equal-FLOP comparisons account for paired branches and candidate scoring. Recursive comparisons also match search opportunities, teacher-call limits, and training-compute ceilings. Realized costs include rejected proposals.

\noindent\textbf{Controlled objective decomposition.}
Table~\ref{tab:relation_components} separates invariant consistency, correct-map fitting, and wrong-map contrast while retaining answer and action SFT. All variants share $H_{\mathrm{abl}}$, verified views, action targets, curriculum, initialization, and source-step budget; compute matching follows Section~\ref{sec:augmentation_controls}. The fixed full-model value of $\gamma$ is used wherever contrast is enabled; setting $\gamma=0$ removes contrast without changing the correct-map term $d_e^+$. A0 matches the loss switches of the main-paper ``w/o Rel.'' row but is evaluated under the fixed-harness protocol.

\par\medskip\noindent\begin{minipage}{\linewidth}
\centering
\captionof{table}{Fine-grained objective variants. Inv. is invariant KL, Map. correct-map fitting, and Ctr. wrong-map contrast. Disabled components are shown by a dash.}
\label{tab:relation_components}
\begingroup\tablelook
\setlength{\tabcolsep}{2pt}
\fontsize{8.5}{10}\selectfont
\begin{tabularx}{\linewidth}{L{0.065\linewidth}|YYYY|L{0.33\linewidth}}
\toprule[0.8pt]
\rowcolor{tblhead}\textbf{ID}&\textbf{SFT}&\textbf{Inv.}&\textbf{Map.}&\textbf{Ctr.}&\textbf{Switches}\\
\midrule[0.45pt]
\rowcolor{tblstripe}A0 & \yes & -- & -- & -- & $\lambda_{\rm inv}=\lambda_{\rm rel}=0$\\
\rowcolor{white}A1 & \yes & \yes & -- & -- & $\lambda_{\rm rel}=0$\\
\rowcolor{tblstripe}A2 & \yes & -- & \yes & -- & $\lambda_{\rm inv}=\gamma=0$\\
\rowcolor{white}A3 & \yes & -- & \yes & \yes & $\lambda_{\rm inv}=0$\\
\rowcolor{tblstripe}A4 & \yes & \yes & \yes & -- & $\gamma=0$\\
\rowcolor{ours}A5 & \yes & \yes & \yes & \yes & Reference\\
\bottomrule[0.8pt]
\end{tabularx}
\endgroup
\end{minipage}\par\medskip

\Needspace{4\baselineskip}
\noindent\textbf{Relation diagnostics.}
For an eligible reference-changing pair $e$, let $\mathcal G_e=\{f_e^\star\}\cup\mathcal F_e^-$ contain the verified map and its predefined wrong-map alternatives. $\widehat f_e=\arg\min_{f\in\mathcal G_e}d_e(f)$ is the selected map under the main-paper discrepancy $d_e(f)=D_{\mathrm{JS}}(p_\theta^g\Vert M_f p_\theta^l)$. Then
\begin{equation}
\mathrm{CMA}=\frac{100}{|\mathcal P|}\sum_{e\in\mathcal P}\mathbf 1[\widehat f_e=f_e^\star].
\end{equation}
Here $\mathcal P$ contains independently verified eligible pairs. Ties within a fixed numerical tolerance are counted as non-agreement, with the tie rate reported separately. CMA is an offline diagnostic over predictive distributions. Wrong-frame answer rate (WFAR) is the percentage of evaluated rollouts on an eligible subset $\mathcal W$ that reuse the answer from the wrong reference frame. Membership in $\mathcal W$ requires distinct correct answers across frames and an unambiguous, predefined error rule. Endpoint accuracy is reported alongside CMA, since two mutually consistent predictions can both be incorrect. Candidate domains, tolerances, and pair eligibility are fixed before evaluating variants; denominators are shared across variants and reported by benchmark.

\par\medskip\noindent\begin{minipage}{\linewidth}
\centering
\captionof{table}{Visual relation supervision under the shared ablation harness $H_{\mathrm{abl}}$. TIR, MME, and BLINK cells contain Average@4 / Pass@4 (\%). CMA and WFAR are equal-benchmark macro percentages over their eligible populations. Table~\ref{tab:relation_counts} gives pair counts; Table~\ref{tab:wfar_populations_review} distinguishes these rates from the pooled common-subset WFAR in Table~\ref{tab:semantic_controls}. Paired differences appear in Table~\ref{tab:primary_ci}.}
\label{tab:relation_results}
\begingroup\tablelook
\setlength{\tabcolsep}{2pt}
\fontsize{8.5}{10}\selectfont
\begin{tabularx}{\linewidth}{L{0.075\linewidth}|YYYYYY}
\toprule[0.8pt]
\rowcolor{tblhead}\textbf{ID}&\textbf{TIR}&\textbf{MME}&\textbf{BLINK}&\textbf{Macro}&\textbf{CMA $\uparrow$}&\textbf{WFAR $\downarrow$}\\
\midrule[0.45pt]
\rowcolor{tblstripe}A0 & \scorepair{23.36}{37.70} & \scorepair{59.17}{74.81} & \scorepair{68.53}{84.62} & \scorepair{50.35}{65.71} & 59.74 & 20.47\\
\rowcolor{white}A1 & \scorepair{24.18}{38.25} & \scorepair{60.28}{76.67} & \scorepair{69.06}{85.31} & \scorepair{51.17}{66.74} & 66.95 & 18.02\\
\rowcolor{tblstripe}A2 & \scorepair{24.45}{38.80} & \scorepair{60.56}{77.04} & \scorepair{69.49}{85.66} & \scorepair{51.50}{67.17} & 74.37 & 15.32\\
\rowcolor{white}A3 & \scorepair{25.27}{39.89} & \scorepair{61.57}{78.15} & \scorepair{70.37}{86.36} & \scorepair{52.40}{68.13} & 79.57 & 12.53\\
\rowcolor{tblstripe}A4 & \scorepair{25.41}{40.44} & \scorepair{61.85}{78.52} & \scorepair{70.72}{86.71} & \scorepair{52.66}{68.56} & 82.89 & 10.66\\
\rowcolor{ours}A5 & \scorepair{25.82}{40.98} & \scorepair{62.69}{79.26} & \scorepair{71.15}{87.41} & \scorepair{53.22}{69.22} & 89.06 & 8.10\\
\bottomrule[0.8pt]
\end{tabularx}
\endgroup
\end{minipage}
\par\medskip

\par\medskip\noindent\begin{minipage}{\linewidth}
\centering
\captionof{table}{Relation coverage and full-objective endpoint/tie diagnostics. Pair counts define the applicable populations for CMA and WFAR.}
\label{tab:relation_counts}
\begingroup\tablelook
\setlength{\tabcolsep}{2pt}
\fontsize{8.5}{10}\selectfont
\begin{tabularx}{\linewidth}{L{0.26\linewidth}|YYYY}
\toprule[0.8pt]
\rowcolor{tblhead}\textbf{Data}&\thcell{CMA\\pairs}&\thcell{WFAR\\pairs}&\thcell{Endpt.\\Avg.@4}&\thcell{Ties\\(\%)}\\
\midrule[0.45pt]
\rowcolor{tblstripe}TIR & 137 & 119 & 66.33 & 2.19\\
\rowcolor{white}MME & 181 & 163 & 75.90 & 1.10\\
\rowcolor{tblstripe}BLINK & 167 & 149 & 82.41 & 2.40\\
\bottomrule[0.8pt]
\end{tabularx}
\endgroup
\end{minipage}\par\medskip

\noindent\textbf{Observed relation effects.}
Across the reported A0--A5 variants, accuracy and correct-map agreement rise monotonically, while wrong-frame answer rate falls (Table~\ref{tab:relation_results}). The full objective (A5) exceeds A0 by 2.87 macro Average@4 and 3.51 Pass@4 points. This macro score balances the reference-changing and remaining test strata within each benchmark, while reference-changing fitting and contrast operate on eligible view pairs. On the common relation-eligible subset, A5 exceeds A0 by 10.93 Average@4 points with 95\% CI $[6.71,15.07]$ (Tables~\ref{tab:semantic_controls} and~\ref{tab:secondary_ci}), resolving a positive effect on the targeted relation tasks. Adding wrong-map contrast (A5 versus A4) changes macro Average@4 by 0.56 points, raises CMA by 6.17 points, and lowers WFAR by 2.56 points. Correct-map fitting (A4 versus A1) adds 1.49 macro Average@4 points, and invariant consistency (A5 versus A3) adds 0.81 points. Table~\ref{tab:primary_ci} reports the macro contrast intervals.

\noindent\textbf{Primary contrasts.}
A5 versus A4 compares wrong-map contrast, A5 versus A3 compares invariant consistency, and A4 versus A1 compares correct-map fitting. Table~\ref{tab:relation_results} reports their accuracy, CMA, and WFAR point estimates; Table~\ref{tab:primary_ci} reports paired accuracy-difference intervals.

\noindent\textbf{Relation behavior.}
The benchmark-macro evaluation measures overall task performance, and the common relation subset measures behavior on the structure targeted by relation supervision. CMA tracks agreement with the verified map, endpoint accuracy tracks answer correctness, and WFAR tracks reuse of an answer from the wrong frame.

\Needspace{4\baselineskip}
\section{Unseen Transformation Generalization}
\label{sec:transformation_generalization}
\noindent\textbf{Split construction.}
Source identities and image hashes keep all derived views within one split. Parameter generalization uses held-out sources and nonoverlapping ranges of the same operations. The crop-footprint split uses $[0.40,0.80]$ for training and $[0.15,0.35]$ for unseen evaluation, where footprint is crop area divided by source-image area. These ranges preserve evidence and are fixed before evaluation. Composition tests train on single operations and evaluate valid crop--resize, resize--crop, rotate--crop, and crop--resize--rotate chains. Their mapping is $f_{\mathrm{combined}}=f_m\circ\cdots\circ f_1$, with compatible intermediate domains and independently verified endpoint answers. The reported conditions concern these held-out magnitudes and compositions of the recorded operations.

\noindent\textbf{Leakage controls and comparators.}
Held-out parameter ranges, operation compositions, and all derived test examples are excluded from teacher training-data generation and acceptance selection. All methods share one training snapshot and the fixed ablation harness $H_{\mathrm{abl}}$. Aug-SFT uses verified augmented views in random order; Curriculum-SFT adds progressive admission; + Invariance adds only the invariant loss; and the full objective adds reference-changing fitting and contrast. Test answers and verified maps are available only to offline scoring, never to the agent. Table~\ref{tab:generalization} reports equally weighted benchmark macro scores.

\par\medskip\noindent\begin{minipage}{\linewidth}
\centering
\captionof{table}{Generalization to held-out transformations under $H_{\mathrm{abl}}$. Each accuracy cell reports Average@4 / Pass@4 (\%). CMA on unseen conditions is reported separately for parameter and composition shifts.}
\label{tab:generalization}
\begingroup\tablelook
\setlength{\tabcolsep}{2pt}
\fontsize{8.5}{10}\selectfont
\begin{tabularx}{\linewidth}{L{0.28\linewidth}|YYYYY}
\toprule[0.8pt]
\rowcolor{tblhead}\textbf{Training}&\thcell{Seen\\$\uparrow$}&\thcell{Param.\\$\uparrow$}&\thcell{Comp.\\$\uparrow$}&\thcell{CMA\\param.}&\thcell{CMA\\comp.}\\
\midrule[0.45pt]
\rowcolor{tblstripe}Aug-SFT & \scorepair{49.09}{64.14} & \scorepair{47.02}{61.19} & \scorepair{44.81}{57.55} & 52.48 & 40.96\\
\rowcolor{white}Curriculum-SFT & \scorepair{50.35}{65.71} & \scorepair{48.59}{62.94} & \scorepair{46.32}{61.43} & 58.97 & 47.04\\
\rowcolor{tblstripe}+ Invariance & \scorepair{51.17}{66.74} & \scorepair{49.92}{65.00} & \scorepair{48.12}{62.77} & 66.39 & 55.71\\
\rowcolor{ours}Full objective & \scorepair{53.22}{69.22} & \scorepair{52.21}{67.95} & \scorepair{51.17}{66.63} & 83.48 & 76.02\\
\bottomrule[0.8pt]
\end{tabularx}
\endgroup
\end{minipage}
\par\medskip

\noindent\textbf{Observed transformation transfer.}
The full objective exceeds Aug-SFT by 5.19 Average@4 points on unseen parameter magnitudes (95\% CI $[1.54,8.94]$) and 6.36 points on unseen compositions ($[2.55,10.16]$; Table~\ref{tab:secondary_ci}). On the composition split, CMA rises from 40.96 to 76.02. The seen-to-composition accuracy drop is 2.05 points for the full objective and 4.28 for Aug-SFT.

\Needspace{4\baselineskip}
\section{Recursive Alternating Optimization}
\label{sec:recursive_controls}
\noindent\textbf{Equal-budget schedules.}
One-pass optimization performs Harness-RSI followed by Model-RSI once; the recursive variants repeat this sequence for two, three, or four rounds. Let $J_{\mathrm{tot}}$ and $U_{\mathrm{tot}}$ be the configured total harness-search steps and optimizer updates. For $r$ rounds, each receives approximately $J_{\mathrm{tot}}/r$ and $U_{\mathrm{tot}}/r$, with integer remainders assigned by a fixed schedule. Candidate limits, total teacher-call caps, training compute, and selection-evaluation opportunities are controlled across schedules; actual consumption is reported rather than inferred from the number of rounds. An equal-update schedule is also audited for realized FLOPs because the active relation pairs can differ across rounds. Budget totals come from the actual run configuration.

\noindent\textbf{Trajectory refresh.}
The three-round fixed-data control reuses the first verified snapshot in later model updates. The frozen-refresh control collects later trajectories with the initial model while retaining the current harness and the common task schedule. These controls appear alongside one-pass and multi-round RSI in Table~\ref{tab:recursive_results}.

\par\medskip\noindent\begin{minipage}{\linewidth}
\centering
\captionof{table}{Alternation under matched total budget ceilings. Macro performance is in percent. FLOPs include all model-training branches; teacher calls include proposal and supervision generation.}
\label{tab:recursive_results}
\begingroup\tablelook
\setlength{\tabcolsep}{2pt}
\fontsize{8.5}{10}\selectfont
\begin{tabularx}{\linewidth}{L{0.29\linewidth}|YYYYY}
\toprule[0.8pt]
\rowcolor{tblhead}\textbf{Schedule}&\thcell{Macro\\Avg.}&\thcell{Macro\\Pass}&\thcell{FLOPs\\($10^{15}$)}&\thcell{Teacher\\calls}&\thcell{Search\\steps}\\
\midrule[0.45pt]
\rowcolor{tblstripe}One-pass & 51.49 & 66.99 & 8.412 & 1297 & 18\\
\rowcolor{white}2-round RSI & 52.66 & 68.56 & 8.397 & 1283 & 18\\
\rowcolor{ours}3-round RSI & 53.61 & 69.64 & 8.431 & 1311 & 18\\
\rowcolor{white}4-round RSI & 53.45 & 69.22 & 8.419 & 1304 & 18\\
\rowcolor{tblstripe}3-round, fixed data & 52.22 & 67.60 & 8.385 & 907 & 18\\
\rowcolor{white}3-round, frozen refresh & 52.62 & 68.44 & 8.426 & 1289 & 18\\
\bottomrule[0.8pt]
\end{tabularx}
\endgroup
\end{minipage}
\par\medskip

\noindent\textbf{Observed recursive endpoints.}
Three-round RSI exceeds one-pass optimization by 2.12 macro Average@4 and 2.65 Pass@4 points, at training costs of $8.431$ versus $8.412\times10^{15}$ FLOPs. Its Average@4 advantages over fixed-data and frozen-model-refresh controls are 1.39 and 0.99 points. Three rounds achieve the highest observed macro Average@4 under these budget ceilings, with four rounds 0.17 points lower (Table~\ref{tab:secondary_ci}). Table~\ref{tab:primary_ci} gives the three-round-minus-one-pass interval $[-1.71,5.92]$; Table~\ref{tab:secondary_ci} gives the schedule and refresh contrasts. Table~\ref{tab:round_history_review} separately records accepted harness updates in all three rounds.

\Needspace{4\baselineskip}
\section{Augmentation-Based SFT Controls}
\label{sec:augmentation_controls}
\noindent\textbf{Same-data controls.}
Raw SFT uses original views only. Aug-SFT adds verified augmented views in random order; Curriculum Aug-SFT adds progressive admission; Curriculum + Invariance adds the invariant loss; and Full Model-RSI uses the complete objective. Except for Raw SFT, all methods share source tasks, augmented images, verified answer and action targets, initialization, LoRA settings, and the fixed harness. Source-step comparisons match exposure and update totals, while measured-FLOP comparisons match training compute. These fixed-harness controls isolate model-learning choices; the end-to-end dual-loop evaluation additionally includes harness evolution.

\par\medskip\noindent\begin{minipage}{\linewidth}
\centering
\captionof{table}{Same-data comparison under $H_{\mathrm{abl}}$. Each benchmark and macro cell reports Average@4 / Pass@4 (\%). Full Model-RSI uses the complete model-update objective.}
\label{tab:augmentation_results}
\begingroup\tablelook
\setlength{\tabcolsep}{2pt}
\fontsize{8.5}{10}\selectfont
\begin{tabularx}{\linewidth}{L{0.34\linewidth}|YYYY}
\toprule[0.8pt]
\rowcolor{tblhead}\textbf{Training}&\textbf{TIR}&\textbf{MME}&\textbf{BLINK}&\textbf{Macro}\\
\midrule[0.45pt]
\rowcolor{tblstripe}Raw SFT & \scorepair{20.63}{33.88} & \scorepair{55.46}{70.74} & \scorepair{65.38}{81.82} & \scorepair{47.16}{62.15}\\
\rowcolor{white}Aug-SFT & \scorepair{22.27}{35.52} & \scorepair{57.87}{73.33} & \scorepair{67.13}{83.57} & \scorepair{49.09}{64.14}\\
\rowcolor{tblstripe}Curriculum SFT & \scorepair{23.36}{37.70} & \scorepair{59.17}{74.81} & \scorepair{68.53}{84.62} & \scorepair{50.35}{65.71}\\
\rowcolor{white}Curriculum + Inv. & \scorepair{24.18}{38.25} & \scorepair{60.28}{76.67} & \scorepair{69.06}{85.31} & \scorepair{51.17}{66.74}\\
\rowcolor{ours}Full Model-RSI & \scorepair{25.82}{40.98} & \scorepair{62.69}{79.26} & \scorepair{71.15}{87.41} & \scorepair{53.22}{69.22}\\
\bottomrule[0.8pt]
\end{tabularx}
\endgroup
\end{minipage}\par\medskip

\par\medskip\noindent\begin{minipage}{\linewidth}
\centering
\captionof{table}{Verified snapshot statistics by benchmark. Acceptance rate is the number of accepted generated views divided by the number proposed; relation-pair counts are reported separately.}
\label{tab:training_statistics}
\begingroup\tablelook
\setlength{\tabcolsep}{2pt}
\fontsize{8.5}{10}\selectfont
\begin{tabularx}{\linewidth}{L{0.42\linewidth}|YYY}
\toprule[0.8pt]
\rowcolor{tblhead}\textbf{Statistic}&\textbf{TIR}&\textbf{MME}&\textbf{BLINK}\\
\midrule[0.45pt]
\rowcolor{tblstripe}Source tasks & 366 & 540 & 572\\
\rowcolor{white}Verified augmented views & 943 & 1375 & 1482\\
\rowcolor{tblstripe}Answer-preserving pairs & 317 & 421 & 389\\
\rowcolor{white}Reference-changing pairs & 249 & 367 & 341\\
\rowcolor{tblstripe}Action-supervision examples & 528 & 763 & 811\\
\rowcolor{white}View acceptance (\%) & 86.75 & 86.42 & 87.54\\
\bottomrule[0.8pt]
\end{tabularx}
\endgroup
\end{minipage}\par\medskip

\par\medskip\noindent\begin{minipage}{\linewidth}
\centering
\captionof{table}{Training controls under $H_{\mathrm{abl}}$. Curric. denotes curriculum and Inv. invariant KL. Augmented-view methods share the verified pool; Raw SFT uses original views. Both budget protocols count paired branches and relation scoring.}
\label{tab:compute_controls}
\begingroup\tablelook
\setlength{\tabcolsep}{2pt}
\fontsize{8}{9.5}\selectfont
\begin{tabularx}{\linewidth}{L{0.31\linewidth}|YYYY}
\toprule[0.65pt]
\rowcolor{tblhead}\textbf{Training}&\thcell{Avg.\\@4}&\thcell{Pass\\@4}&\textbf{Updates}&\thcell{Tokens\\(M)}\\
\midrule[0.4pt]
\rowcolor{tblhead}\multicolumn{5}{l}{\textit{Equal source steps}}\\
\rowcolor{white}Raw SFT & 47.16 & 62.15 & 1800 & 2.917\\
\rowcolor{tblstripe}Aug-SFT & 49.09 & 64.14 & 1800 & 4.683\\
\rowcolor{white}Curric. SFT & 50.35 & 65.71 & 1800 & 4.512\\
\rowcolor{tblstripe}Curric. + Inv. & 51.17 & 66.74 & 1800 & 5.791\\
\rowcolor{ours}Full Model-RSI & 53.22 & 69.22 & 1800 & 7.238\\
\rowcolor{tblhead}\multicolumn{5}{l}{\textit{Equal training FLOPs}}\\
\rowcolor{white}Raw SFT & 47.70 & 62.39 & 4768 & 7.727\\
\rowcolor{tblstripe}Aug-SFT & 49.32 & 64.26 & 3125 & 8.130\\
\rowcolor{white}Curric. SFT & 50.61 & 65.84 & 3209 & 8.044\\
\rowcolor{tblstripe}Curric. + Inv. & 51.31 & 66.98 & 2377 & 7.647\\
\rowcolor{ours}Full Model-RSI & 53.22 & 69.22 & 1800 & 7.238\\
\bottomrule[0.65pt]
\end{tabularx}

\par\smallskip
\begin{tabularx}{\linewidth}{L{0.31\linewidth}|YYYY}
\toprule[0.65pt]
\rowcolor{tblhead}\textbf{Training}&\thcell{Forward\\passes}&\thcell{FLOPs\\($10^{15}$)}&\thcell{Time\\(h)}&\thcell{Peak\\GB}\\
\midrule[0.4pt]
\rowcolor{tblhead}\multicolumn{5}{l}{\textit{Equal source steps}}\\
\rowcolor{white}Raw SFT & 35424 & 3.183 & 1.87 & 23.74\\
\rowcolor{tblstripe}Aug-SFT & 55008 & 4.856 & 2.73 & 28.91\\
\rowcolor{white}Curric. SFT & 53856 & 4.729 & 2.68 & 29.08\\
\rowcolor{tblstripe}Curric. + Inv. & 75168 & 6.384 & 3.51 & 37.43\\
\rowcolor{ours}Full Model-RSI & 98784 & 8.431 & 4.62 & 48.62\\
\rowcolor{tblhead}\multicolumn{5}{l}{\textit{Equal training FLOPs}}\\
\rowcolor{white}Raw SFT & 93834 & 8.431 & 4.92$^\dagger$ & 23.74\\
\rowcolor{tblstripe}Aug-SFT & 95500 & 8.431 & 4.71 & 28.91\\
\rowcolor{white}Curric. SFT & 96013 & 8.431 & 4.63 & 29.08\\
\rowcolor{tblstripe}Curric. + Inv. & 99264 & 8.431 & 4.74 & 37.43\\
\rowcolor{ours}Full Model-RSI & 98784 & 8.431 & 4.62 & 48.62\\
\bottomrule[0.65pt]
\end{tabularx}
\par\smallskip\raggedright\footnotesize $^\dagger$Estimated as 0.02 h setup + per-update time calibrated to the 1,800-update run.\par
\endgroup
\end{minipage}
\par\medskip
\noindent\textbf{Measured training-control comparison.}
Full Model-RSI exceeds Aug-SFT by 4.13 Average@4 and 5.08 Pass@4 points at equal source steps. At the common measured budget of $8.431\times10^{15}$ training FLOPs, the differences are 3.90 and 4.96 points. The paired Average@4 intervals are $[0.31,7.92]$ and $[0.15,7.65]$ points, respectively (Table~\ref{tab:primary_ci}), supporting a benefit at both equal source exposure and equal training compute.

\Needspace{4\baselineskip}
\section{Harness Internalization and Efficiency}
\label{sec:internalization}
\noindent\textbf{Separating parameter gains from external guidance.}
Let $\theta_0$ and $\theta_T$ denote the initial and final models, $H_0$ the initial harness, $H_T^+$ the final uncompressed harness, and $\widehat H_T$ its compacted version. A minimal compatible harness $H_{\min}$ retains the same tool APIs and basic execution constraints while removing learned task-specific guidance. Table~\ref{tab:internalization} crosses these components on the same held-out tasks. Distinct from the selection objective $A$, $a(\theta,H)$ denotes held-out Average@4. We quantify parameter gains and dependence on external guidance with
\begin{equation}
G_{\mathrm{int}}=a(\theta_T,H_0)-a(\theta_0,H_0),
\end{equation}
\begin{equation}
D(\theta)=a(\theta,H_T^+)-a(\theta,H_{\min}).
\end{equation}
\par\medskip\noindent\begin{minipage}{\linewidth}
\centering
\captionof{table}{Crossed model--harness evaluation for reference R0; cells show Average@4 / Pass@4 (\%).}
\label{tab:internalization}
\begingroup\tablelook
\setlength{\tabcolsep}{2pt}
\fontsize{8.5}{10}\selectfont
\begin{tabularx}{\linewidth}{L{0.10\linewidth}|L{0.25\linewidth}|YYYY}
\toprule[0.8pt]
\rowcolor{tblhead}\textbf{Model}&\textbf{Harness}&\textbf{TIR}&\textbf{MME}&\textbf{BLINK}&\textbf{Macro}\\
\midrule[0.45pt]
\rowcolor{tblstripe}$\theta_0$ & Initial $H_0$ & \scorepair{18.44}{29.51} & \scorepair{53.89}{69.26} & \scorepair{62.15}{79.72} & \scorepair{44.83}{59.50}\\
\rowcolor{white}$\theta_T$ & Initial $H_0$ & \scorepair{24.04}{38.25} & \scorepair{60.37}{77.04} & \scorepair{68.18}{85.31} & \scorepair{50.87}{66.87}\\
\rowcolor{tblstripe}$\theta_T$ & Full $H_T^+$ & \scorepair{26.09}{41.53} & \scorepair{63.33}{79.63} & \scorepair{71.42}{87.76} & \scorepair{53.61}{69.64}\\
\rowcolor{ours}$\theta_T$ & Compact $\widehat H_T$ & \scorepair{26.09}{41.53} & \scorepair{63.33}{79.63} & \scorepair{71.42}{87.76} & \scorepair{53.61}{69.64}\\
\rowcolor{tblstripe}$\theta_T$ & Minimal $H_{\min}$ & \scorepair{23.36}{36.61} & \scorepair{59.54}{75.93} & \scorepair{67.48}{84.27} & \scorepair{50.13}{65.60}\\
\rowcolor{white}$\theta_0$ & Full $H_T^+$ & \scorepair{21.58}{33.33} & \scorepair{55.28}{67.78} & \scorepair{64.51}{80.07} & \scorepair{47.12}{60.39}\\
\rowcolor{tblstripe}$\theta_0$ & Minimal $H_{\min}$ & \scorepair{15.71}{25.68} & \scorepair{50.83}{65.19} & \scorepair{59.62}{77.27} & \scorepair{42.05}{56.05}\\
\rowcolor{white}Aug-SFT & Initial $H_0$ & \scorepair{21.31}{33.88} & \scorepair{56.94}{73.33} & \scorepair{65.21}{82.52} & \scorepair{47.82}{63.24}\\
\rowcolor{tblstripe}Aug-SFT & Ablation $H_{\mathrm{abl}}$ & \scorepair{22.27}{35.52} & \scorepair{57.87}{73.33} & \scorepair{67.13}{83.57} & \scorepair{49.09}{64.14}\\
\rowcolor{white}Aug-SFT & Full $H_T^+$ & \scorepair{23.63}{37.70} & \scorepair{59.63}{75.56} & \scorepair{68.09}{84.97} & \scorepair{50.45}{66.08}\\
\rowcolor{tblstripe}Aug-SFT & Minimal $H_{\min}$ & \scorepair{20.49}{32.24} & \scorepair{55.56}{71.85} & \scorepair{64.42}{81.82} & \scorepair{46.82}{61.97}\\
\bottomrule[0.8pt]
\end{tabularx}
\endgroup
\end{minipage}\par\medskip
$G_{\mathrm{int}}$ measures parameter-associated gains under unchanged external guidance. The full and minimal harnesses retain compatible tool interfaces for both models, allowing $D(\theta_0)$ and $D(\theta_T)$ to quantify the guidance gap before and after learning.

\noindent\textbf{Observed crossed-harness effects.}
The macro values give $G_{\rm int}=6.04$ percentage points, $D(\theta_0)=5.07$, and $D(\theta_T)=3.48$. The full-versus-minimal guidance gap decreases by 1.59 points while minimal-harness accuracy rises from 42.05 to 50.13. Under $H_0$, the final model exceeds same-data Aug-SFT by 3.05 points; under $H_{\min}$, the difference is 3.31 points. Full and compact guidance yield the same reported accuracy, with online costs in Table~\ref{tab:online_efficiency}.

\noindent\textbf{Matched-data training control.}
The Aug-SFT checkpoint in Table~\ref{tab:internalization} is identical to the Aug-SFT checkpoint in Table~\ref{tab:augmentation_results}. Its macro Average@4 scores under $H_0$, $H_{\mathrm{abl}}$, and $H_T^+$ are 47.82, 49.09, and 50.45, respectively; only the evaluation harness changes. The full-objective checkpoint scores higher under the shared initial, full, and minimal harnesses.

\noindent\textbf{Online efficiency and compaction.}
Using a fixed tokenizer, harness compression is $\mathrm{CR}=1-T(\widehat H_T)/T(H_T^+)$. Online measurements use matched hardware, image processing, decoding limits, concurrency, and cache policy, including failed attempts. Costs average the four rollouts within each task, then tasks within each benchmark, and finally benchmarks with equal weights. They are per-rollout means, rather than four-rollout sums. Latency is the mean end-to-end time of one rollout. Input columns in Tables~\ref{tab:online_efficiency} and~\ref{tab:mechanism} count logical unique input $U$: the harness prefix once per rollout and later appended input material when it first appears. Section~\ref{sec:accounting_review} distinguishes $U$ from cumulative full-prompt input $I$, output tokens, and physical cache work. Compaction proposals are included even when rejected.

\par\medskip\noindent\begin{minipage}{\linewidth}
\centering
\captionof{table}{Online cost and benchmark-macro performance for reference R0. Costs are per-rollout means: average four rollouts per task, then tasks and equally weighted benchmarks. Input is logical unique input $U$; full-prompt input $I$ is reported separately in Table~\ref{tab:token_ledger_review}.}
\label{tab:online_efficiency}
\begingroup\tablelook
\setlength{\tabcolsep}{2pt}
\fontsize{8}{9.5}\selectfont
\begin{tabularx}{\linewidth}{L{0.34\linewidth}|YYY}
\toprule[0.65pt]
\rowcolor{tblhead}\textbf{Configuration}&\thcell{Avg. /\\Pass}&\thcell{Harness\\tokens}&\thcell{Unique input\\tokens}\\
\midrule[0.4pt]
\rowcolor{tblstripe}Initial & \scorepair{44.83}{59.50} & 1847 & 13682.40\\
\rowcolor{white}Final, uncompressed & \scorepair{53.61}{69.64} & 4289 & 15937.80\\
\rowcolor{ours}Compact final & \scorepair{53.61}{69.64} & 2263 & 13218.60\\
\bottomrule[0.65pt]
\end{tabularx}

\par\smallskip
\begin{tabularx}{\linewidth}{L{0.34\linewidth}|YYY}
\toprule[0.65pt]
\rowcolor{tblhead}\textbf{Configuration}&\thcell{Output\\tokens}&\thcell{Tool\\calls}&\thcell{Latency\\(s)}\\
\midrule[0.4pt]
\rowcolor{tblstripe}Initial & 1183.70 & 3.18 & 42.73\\
\rowcolor{white}Final, uncompressed & 1057.40 & 2.91 & 39.86\\
\rowcolor{ours}Compact final & 1041.90 & 2.84 & 35.94\\
\bottomrule[0.65pt]
\end{tabularx}
\endgroup
\end{minipage}\par\medskip

\par\medskip\noindent\begin{minipage}{\linewidth}
\centering
\captionof{table}{Cumulative parameter transition under the final harness on the fixed 183-task TIR diagnostic subset, with terminal compaction diagnostics for R0. M0/M1 use the initial/final-model states under $H_T^+$ in Table~\ref{tab:internalization}. This table reports subset accuracy and rollout counts; the crossed table reports stratum-balanced accuracy. Fix/Reg. count wrong-to-correct/correct-to-wrong rollouts; $\Delta$Avg. is in percentage points. The model comparison covers the complete optimization sequence.}
\label{tab:mechanism}
\begingroup\tablelook
\setlength{\tabcolsep}{2pt}
\fontsize{8}{9.5}\selectfont
\begin{tabularx}{\linewidth}{L{0.12\linewidth}|YYYY}
\toprule[0.65pt]
\rowcolor{tblhead}\textbf{State}&\textbf{Avg.@4}&\textbf{$\Delta$Avg.}&\textbf{Fix}&\textbf{Reg.}\\
\midrule[0.4pt]
\rowcolor{tblhead}\multicolumn{5}{l}{\textit{Cumulative parameter change; reference M0}}\\
\rowcolor{white}M0 & 21.58 & Ref. & Ref. & Ref.\\
\rowcolor{tblstripe}M1 & 26.09 & 4.51 & 159 & 126\\
\rowcolor{tblhead}\multicolumn{5}{l}{\textit{Compaction; reference C0}}\\
\rowcolor{tblstripe}C0 & 26.09 & Ref. & Ref. & Ref.\\
\rowcolor{white}C1 & 25.82 & -0.27 & 5 & 7\\
\rowcolor{ours}C2 & 26.09 & 0.00 & 4 & 4\\
\bottomrule[0.65pt]
\end{tabularx}

\par\smallskip
\begin{tabularx}{\linewidth}{L{0.12\linewidth}|YYYY}
\toprule[0.65pt]
\rowcolor{tblhead}\textbf{State}&\thcell{Harness\\tokens}&\thcell{Calls\\M / T}&\thcell{Unique input\\(k)}&\thcell{Time\\(s)}\\
\midrule[0.4pt]
\rowcolor{tblstripe}M0 & 4289 & 4.13 / 3.07 & 16.827 & 41.38\\
\rowcolor{white}M1 & 4289 & 3.87 / 2.86 & 15.904 & 39.27\\
\rowcolor{tblstripe}C0 & 4289 & 3.87 / 2.86 & 15.904 & 39.27\\
\rowcolor{white}C1 & 2817 & 3.79 / 2.81 & 14.198 & 37.62\\
\rowcolor{ours}C2 & 2263 & 3.74 / 2.78 & 13.587 & 35.21\\
\bottomrule[0.65pt]
\end{tabularx}

\par\smallskip\raggedright\footnotesize M0: $(\theta_0,H_T^+)$; M1: $(\theta_T,H_T^+)$. C0 equals M1; C1/C2 apply exact/teacher compaction at fixed $\theta_T$. Table~\ref{tab:benchmark_flips_review} gives terminal-compaction flips for all benchmarks. M/T denote model/tool calls. Unique input excludes the output-token total and is averaged per rollout on TIR only. Ref. marks a comparison reference.\par
\endgroup
\end{minipage}
\par\medskip
\noindent\textbf{Measured deployment and compaction effects.}
Compaction reduces static harness length from 4,289 to 2,263 tokens ($\mathrm{CR}=47.24\%$), while macro accuracy remains 53.61 / 69.64. In Table~\ref{tab:online_efficiency}, unique input decreases from 15,937.80 to 13,218.60 tokens and mean rollout latency from 39.86 to 35.94 seconds (9.83\%). On the 183-task TIR diagnostic subset, the cumulative parameter comparison repairs 159 rollouts and regresses on 126, giving a net 33 out of $183\times4=732$, or 4.51 points within that subset. Exact compaction changes subset accuracy by $-0.27$ points; teacher compaction has zero net subset change with four fixes and four regressions. Table~\ref{tab:secondary_ci} reports the paired compaction intervals.

\noindent\textbf{Offline improvement cost.}
Table~\ref{tab:offline_cost} assigns each call to one phase and includes failed or rejected attempts. GPU-hours sum device-active time; elapsed wall time is measured directly because phases may overlap. Full-run totals are 1,311 teacher calls, 33,980 student rollouts, 55.37 GPU-hours, and 23.78 wall hours. Offline phase-token totals and per-rollout online ledgers are separate accounting quantities.

\par\medskip\noindent\begin{minipage}{\linewidth}
\centering
\captionof{table}{Offline totals include rejected and failed attempts. Each call belongs to one phase. Table~\ref{tab:event_ledger_review} separates harness-incumbent and compaction evaluation events. Tokens are in millions; GPU-hours sum device-active time, and wall time is measured directly. N/A denotes an inapplicable field.}
\label{tab:offline_cost}
\begingroup\tablelook
\setlength{\tabcolsep}{2pt}
\fontsize{8}{9.5}\selectfont
\begin{tabularx}{\linewidth}{L{0.34\linewidth}|YYYY}
\toprule[0.65pt]
\rowcolor{tblhead}\textbf{Phase}&\thcell{Teacher\\calls}&\thcell{Student\\rollouts}&\thcell{Teacher\\tokens}&\thcell{Student\\tokens}\\
\midrule[0.4pt]
\rowcolor{tblstripe}Harness proposals & 54 & N/A & 0.417 & N/A\\
\rowcolor{white}TRAIN collection & N/A & 4376 & N/A & 17.832\\
\rowcolor{tblstripe}Data generation / audit & 1254 & N/A & 6.283 & N/A\\
\rowcolor{white}Harness selection & N/A & 17496 & N/A & 71.284\\
\rowcolor{tblstripe}LoRA training & N/A & N/A & N/A & N/A\\
\rowcolor{white}Model selection & N/A & 8868 & N/A & 36.071\\
\rowcolor{tblstripe}Compaction & 3 & 3240 & 0.064 & 12.947\\
\rowcolor{ours}Total offline & 1311 & 33980 & 6.764 & 138.134\\
\bottomrule[0.65pt]
\end{tabularx}

\par\smallskip
\begin{tabularx}{\linewidth}{L{0.48\linewidth}|YY}
\toprule[0.65pt]
\rowcolor{tblhead}\textbf{Phase}&\textbf{GPU-hours}&\textbf{Wall hours}\\
\midrule[0.4pt]
\rowcolor{tblstripe}Harness proposals & 0.00 & 0.73\\
\rowcolor{white}TRAIN collection & 4.81 & 1.37\\
\rowcolor{tblstripe}Data generation / audit & 0.00 & 8.42\\
\rowcolor{white}Harness selection & 18.93 & 4.97\\
\rowcolor{tblstripe}LoRA training & 18.48 & 4.62\\
\rowcolor{white}Model selection & 9.64 & 2.53\\
\rowcolor{tblstripe}Compaction & 3.51 & 1.14\\
\rowcolor{ours}Total offline & 55.37 & 23.78\\
\bottomrule[0.65pt]
\end{tabularx}
\endgroup
\end{minipage}\par\medskip
\Needspace{4\baselineskip}
\section{Robustness and Sensitivity Diagnostics}
\label{sec:additional_tables}
Category-level changes test whether aggregate gains conceal regressions. Harness ablations separate instruction edits, file edits, history search, and joint review under fixed proposal budgets. Sensitivity tests vary one factor at a time. Candidate domains and defaults are fixed before evaluation.

\par\medskip\noindent\begin{minipage}{\linewidth}
\centering
\captionof{table}{Category-level performance changes for Qwen3-VL-8B reference R0. Scores are Average@4 (\%); changes are in percentage points. Count is the number of task categories, and Worst $\Delta$ is the lowest category-level change.}
\label{tab:retention}
\begingroup\tablelook
\setlength{\tabcolsep}{2pt}
\fontsize{8}{9.5}\selectfont
\setlength{\tabcolsep}{2pt}
\begin{tabularx}{\linewidth}{L{0.30\linewidth}||YY|Y|YY}
\toprule[0.8pt]
\rowcolor{tblhead}\textbf{Benchmark / Student}&\multicolumn{2}{c|}{\textbf{Score $\uparrow$}}&\textbf{Change}&\multicolumn{2}{c}{\textbf{Categories}}\\
\rowcolor{tblhead}&Initial&Final&$\Delta$ score $\uparrow$&Count&Worst $\Delta$\\
\midrule[0.45pt]
\midrule[0.35pt]\rowcolor{tblhead}\multicolumn{6}{l}{\textit{Task categories, Qwen3-VL-8B}}\\
\rowcolor{white}TIR & 18.44 & 26.09 & 7.65 & 9 & -0.82\\
\rowcolor{tblstripe}MME & 53.89 & 63.33 & 9.44 & 11 & -0.37\\
\rowcolor{white}BLINK & 62.15 & 71.42 & 9.27 & 13 & -0.69\\
\bottomrule[0.65pt]
\end{tabularx}
\endgroup
\end{minipage}\par\medskip
\noindent\textbf{Observed category regressions.}
Benchmark Average@4 improves by 7.65, 9.44, and 9.27 points on TIR, MME, and BLINK, respectively. The worst category-level changes are $-0.82$, $-0.37$, and $-0.69$ points.

\par\medskip\noindent\begin{minipage}{\linewidth}
\centering
\captionsetup{singlelinecheck=false,hypcap=false}
\captionof{table}{Harness component ablation on TIR-Bench with the initial Qwen3-VL-8B model $\theta_0$ fixed and matched proposal budgets; the full-control evaluation is shared with the 3-slot row in Table~\ref{tab:sensitivity}(a).}
\label{tab:harness_controls}
\begingroup\tablelook
\setlength{\tabcolsep}{2pt}
\fontsize{8.5}{10}\selectfont
\setlength{\tabcolsep}{2pt}
\renewcommand{\arraystretch}{1.24}
\begin{tabularx}{\linewidth}{>{\centering\arraybackslash}p{0.065\linewidth}|YYYY|YY}
\toprule[0.8pt]
\rowcolor{tblhead}\textbf{ID}&\multicolumn{4}{c|}{\textbf{Harness components}}&\multicolumn{2}{c}{\textbf{Test (\%) $\uparrow$}}\\
\rowcolor{tblhead}&Instr.&Files&History&Review&Avg.@4&Pass@4\\
\midrule[0.45pt]
\rowcolor{tblstripe}1 & \yes & -- & \yes & \yes & 20.63 & 31.69\\
\rowcolor{white}2 & -- & \yes & \yes & \yes & 19.81 & 30.60\\
\rowcolor{tblstripe}3 & \yes & \yes & -- & \yes & 20.08 & 31.15\\
\rowcolor{white}4 & \yes & \yes & \yes & -- & 20.90 & 32.24\\
\rowcolor{ours}\textbf{5} & \yes & \yes & \yes & \yes & 21.58 & 33.33\\
\bottomrule[0.8pt]
\end{tabularx}
\par\vspace{4pt}
\raggedright\footnotesize
Instr./Files denote instruction/file edits; History/Review denote history search/joint review. A dash disables a component. Disabling history search uses the incumbent only; disabling joint review reviews trajectories separately. Row 5 enables all components. Row 5 and the $\theta_0/H_T^+$ row in Table~\ref{tab:internalization} give equal balanced TIR scores; they are distinct harnesses (3,041 vs. 4,289 tokens).\par
\endgroup
\end{minipage}\par\medskip
\par\medskip\noindent\begin{minipage}{\linewidth}
\centering
\captionof{table}{Search-budget and relation-objective sensitivity. Panel (a) compares separate six-batch searches with $\theta_0$ fixed and no model updates; its 3-slot condition reuses the 3,041-token full fixed-model control in Table~\ref{tab:harness_controls}. Panel (b) varies one factor around reference R0 while keeping others fixed. Caps are configured limits; defaults, candidate domains, and bins are fixed before test evaluation.}
\label{tab:sensitivity}
\begingroup\tablelook
\setlength{\tabcolsep}{2pt}
\fontsize{8}{9.5}\selectfont
\setlength{\tabcolsep}{2pt}
\begin{tabularx}{\linewidth}{L{0.27\linewidth}||YY|YYYY}
\toprule[0.8pt]
\rowcolor{tblhead}\textbf{Setting}&\multicolumn{2}{c|}{\textbf{Test (\%) $\uparrow$}}&\multicolumn{4}{c}{\textbf{Search accounting}}\\
\rowcolor{tblhead}&Avg.&Pass&Cap&Unique&Invalid&Dup.\\
\midrule[0.45pt]
\midrule[0.35pt]\rowcolor{tblhead}\multicolumn{7}{l}{\textit{(a) Harness proposal budget}}\\
\rowcolor{white}1 slot / batch & 19.54 & 30.60 & 6 & 4 & 1 & 1\\
\rowcolor{tblstripe}2 slots / batch & 20.77 & 32.24 & 12 & 9 & 2 & 1\\
\rowcolor{tblstripe}3 slots / batch & 21.58 & 33.33 & 18 & 14 & 3 & 1\\
\midrule[0.35pt]\rowcolor{tblhead}\multicolumn{7}{l}{\textit{(b) Relation objective: Avg.@4 / Pass@4 (\%)}}\\
\rowcolor{white}\textbf{Factor} & \multicolumn{2}{c|}{\textbf{Lower}} & \multicolumn{2}{c}{\textbf{Default}} & \multicolumn{2}{c}{\textbf{Higher}}\\
\rowcolor{tblstripe}Loss weight $\lambda_{\mathrm{rel}}$ & \multicolumn{2}{c|}{\shortstack[c]{$x_0/2$\\25.55\,/\,40.44}} & \multicolumn{2}{c}{\shortstack[c]{$x_0$\\26.09\,/\,41.53}} & \multicolumn{2}{c}{\shortstack[c]{$2x_0$\\25.82\,/\,40.98}}\\
\rowcolor{white}Relation temperature & \multicolumn{2}{c|}{\shortstack[c]{$x_0/2$\\25.68\,/\,40.98}} & \multicolumn{2}{c}{\shortstack[c]{$x_0$\\26.09\,/\,41.53}} & \multicolumn{2}{c}{\shortstack[c]{$2x_0$\\25.27\,/\,40.44}}\\
\rowcolor{tblstripe}Candidates / bins & \multicolumn{2}{c|}{\shortstack[c]{$x_0/2$\\25.41\,/\,40.44}} & \multicolumn{2}{c}{\shortstack[c]{$x_0$\\26.09\,/\,41.53}} & \multicolumn{2}{c}{\shortstack[c]{$2x_0$\\25.96\,/\,41.53}}\\
\rowcolor{white}Wrong-map count & \multicolumn{2}{c|}{\shortstack[c]{$x_0/2$\\25.14\,/\,39.89}} & \multicolumn{2}{c}{\shortstack[c]{$x_0$\\26.09\,/\,41.53}} & \multicolumn{2}{c}{\shortstack[c]{$2x_0$\\25.82\,/\,41.53}}\\
\bottomrule[0.8pt]
\end{tabularx}
\endgroup

\end{minipage}\par\medskip

\section{Semantic Structure and Explicit Conversion}
\label{sec:semantic_controls}
\noindent\textbf{Comparison scope.}
Table~\ref{tab:semantic_controls} separates two comparisons. Panel (a) contrasts correct relations with no relation supervision and corrupted auxiliary relations. Panel (b) compares direct answers with deterministic conversion on a shared conversion-eligible subset. These panels concern different evaluation populations.

\noindent\textbf{Controls.}
The relation controls share a fixed harness, verified snapshot, source tasks, answer/action targets, curriculum, initialization, and LoRA settings. The no-relation and correct-relation configurations correspond to A0 and A5. Within-type permutation changes the auxiliary pairing between compatible reference-changing views while keeping individual SFT contexts and labels intact. The incorrect-map control retains the original view pairs and changes the positive reference-changing map. Invariant supervision is unchanged in both corruption controls. Panel (a) reports the resulting scores on the common relation-evaluation subset.

\noindent\textbf{Deterministic conversion.}
The conversion comparator applies a deterministic program to a local prediction from the common-pool Aug-SFT model, using recorded operation parameters to express the answer in the requested frame. Conversion is restricted to tasks for which deployment-available metadata defines the mapping, such as valid area and coordinate transformations. It does not use verified test answers, gold target locations, or answer-derived mappings. Crop counts are excluded because they do not generally determine full-image counts without evidence outside the crop. Panel (b) also applies the converter to the full-objective checkpoint, separating the reported direct-prediction scores from the added conversion step.

\par\medskip\noindent\begin{minipage}{\linewidth}
\centering
\captionof{table}{Semantic-structure controls and explicit conversion. Avg. is endpoint-averaged held-out Average@4; JCA and WFAR are defined in Section~\ref{sec:behavioral_diagnostics}. Panel (a) pools 151 source pairs across benchmarks, whereas Table~\ref{tab:relation_results} macro-averages benchmark-specific WFAR. Panel (b) uses a distinct 113-pair conversion-eligible subset. Coverage and WFAR denominators are in Table~\ref{tab:wfar_populations_review}; paired differences are in Table~\ref{tab:secondary_ci}.}
\label{tab:semantic_controls}
\begingroup\tablelook\fontsize{8}{9.5}\selectfont
\begin{tabularx}{\linewidth}{L{0.43\linewidth}|YYY}
\toprule[0.65pt]
\rowcolor{tblhead}\textbf{Training / inference}&\textbf{Avg.}&\textbf{JCA}&\textbf{WFAR}\\
\midrule[0.4pt]
\rowcolor{tblhead}\multicolumn{4}{l}{\textit{(a) Common relation-evaluation subset}}\\
\rowcolor{white}No relation (A0)&62.17&43.87&17.22\\
\rowcolor{tblstripe}Permuted reference pairs&61.51&41.23&19.04\\
\rowcolor{white}Incorrect positive maps&60.51&39.74&21.36\\
\rowcolor{ours}Correct relations (A5)&73.10&59.27&8.61\\
\midrule[0.35pt]
\rowcolor{tblhead}\multicolumn{4}{l}{\textit{(b) Shared conversion-eligible subset}}\\
\rowcolor{white}Aug-SFT, direct answer&64.49&47.12&15.27\\
\rowcolor{tblstripe}Aug-SFT + converter&70.91&55.53&9.73\\
\rowcolor{white}Full objective, direct answer&74.67&60.84&7.30\\
\rowcolor{ours}Full objective + converter&74.89&63.27&6.19\\
\bottomrule[0.65pt]
\end{tabularx}\endgroup
\end{minipage}\par\medskip

\noindent\textbf{Observed semantic and conversion effects.}
On the common relation subset, correct relations exceed A0 by 10.93 Average@4 points (95\% CI $[6.71,15.07]$), raise JCA by 15.40 points, and lower WFAR by 8.61 points. Both corrupted-relation controls score below A0, connecting the gain to the correct relation semantics. On the conversion-eligible subset, adding the converter improves Aug-SFT by 6.42 accuracy points. The full objective\textquotesingle s direct answer exceeds Aug-SFT with conversion by 3.76 points, and adding conversion to the full objective changes accuracy by 0.22 points (CI $[-4.65,4.87]$; Table~\ref{tab:secondary_ci}).

\noindent\textbf{Relation semantics and conversion.}
The relation-eligible evaluation focuses on tasks where the recorded operation links answers across reference frames. Explicit conversion yields a larger point-estimate gain for Aug-SFT, while the full objective already achieves higher direct-answer accuracy. Table~\ref{tab:wfar_populations_review} specifies the populations and weighting used for these comparisons.

\section{Independent Supervision Audit}
\label{sec:supervision_audit}
\noindent\textbf{Audit populations.}
Table~\ref{tab:supervision_audit} distinguishes source tasks from derived records. The accepted-view checks and raw annotator agreement cover 281 records from 137 sources. Relation validity covers 227 records from 113 sources, and the rejected-record audit covers 127 records from 79 sources.

The table reports evidence sufficiency, answer correctness, relation validity, the fraction of rejected records judged valid, and raw annotator agreement. Each rate is shown with its reported uncertainty interval. The aggregate rates summarize supervision quality in these audited samples, with source and record counts specifying their coverage.

\par\medskip\noindent\begin{minipage}{\linewidth}
\centering
\captionof{table}{Independent audit of generated supervision. Each rate includes an uncertainty interval. ``Sources'' counts distinct source tasks and ``Records'' counts audited derived examples. The table summarizes aggregate audited counts, rates, and intervals.}
\label{tab:supervision_audit}
\begingroup\tablelook\fontsize{8}{9.5}\selectfont
\begin{tabularx}{\linewidth}{L{0.38\linewidth}|YYY}
\toprule[0.65pt]
\rowcolor{tblhead}\textbf{Audit target}&\textbf{Sources}&\textbf{Records}&\textbf{Rate (\%)}\\
\midrule[0.4pt]
\rowcolor{white}Accepted: sufficient evidence&137&281&\shortstack[c]{96.80\\{[94.3,98.9]}}\\
\rowcolor{tblstripe}Accepted: correct answer&137&281&\shortstack[c]{94.31\\{[91.5,96.8]}}\\
\rowcolor{white}Accepted: valid relation&113&227&\shortstack[c]{92.07\\{[88.1,95.6]}}\\
\rowcolor{tblstripe}Rejected: actually valid&79&127&\shortstack[c]{14.17\\{[8.5,20.3]}}\\
\rowcolor{white}Raw annotator agreement&137&281&\shortstack[c]{91.81\\{[88.5,94.9]}}\\
\bottomrule[0.65pt]
\end{tabularx}\endgroup
\end{minipage}\par\medskip

\section{Free-Generation Relation Metrics}
\label{sec:behavioral_diagnostics}
\noindent\textbf{Metric definition.}
CMA scores candidate distributions, whereas JCA measures joint correctness of independently generated answers for both endpoints. The endpoint requests specify their answer frames without exposing the candidate list or verified mapping. For eligible pair $e$ and rollout seed $s$, correctness indicators $z_{e,s}^l,z_{e,s}^g$ define
\begin{equation}
\mathrm{JCA}=\frac{100}{4|\mathcal P|}
\sum_{e\in\mathcal P}\sum_{s=1}^{4}z_{e,s}^l z_{e,s}^g.
\end{equation}
The four endpoint rollouts use a fixed pairing. Table~\ref{tab:semantic_controls} reports JCA and WFAR for its two evaluation subsets. WFAR identifies free-generated answers that reuse the wrong reference frame on pairs with distinct frame-specific correct answers. Unparseable or invalid answers count as incorrect in accuracy. High CMA alone can coexist with low JCA because relationally compatible predictions can both be wrong.

\section{Paired Differences and Confidence Intervals}
\label{sec:paired_ci}
\noindent\textbf{Computation.}
Tables~\ref{tab:primary_ci} and~\ref{tab:secondary_ci} report percentage-point differences with percentile 95\% intervals from 10,000 paired source-task bootstrap resamples. For stratum-balanced accuracy, source tasks are resampled separately within Test A and Test B of each benchmark, preserving the stratum sizes and all four rollouts per task; the same sampled indices are applied to both methods. Each replicate computes the stratum metrics, averages them with equal weight, and averages the three benchmarks with equal weight. Four-run comparisons share sampled indices across runs and average the macro differences over the four runs. Single-checkpoint ablations condition on one checkpoint per configuration. Compaction contrasts resample the TIR diagnostic subset; semantic comparisons resample source pairs with both endpoints and four rollouts (151 pairs for the relation subset, 113 for conversion). Differences use unrounded metrics and may differ from subtracting displayed cells by 0.01.

\noindent\textbf{Interpretation.}
Intervals are descriptive, unadjusted for multiple comparisons, and conditional on the evaluated checkpoints. The four-run panel is fixed during resampling; sample SD in Tables~\ref{tab:main_multirun} and~\ref{tab:perrun_results} summarizes run-to-run dispersion separately.

\par\medskip\noindent\begin{minipage}{\linewidth}
\centering
\captionof{table}{Primary paired comparisons for Tables~\ref{tab:main_multirun}, \ref{tab:relation_results}, \ref{tab:compute_controls}, and~\ref{tab:recursive_results}. Values are percentage-point differences in stratum-balanced macro accuracy; brackets are 95\% paired, stratified source-task bootstrap intervals, distinct from across-run SD.}
\label{tab:primary_ci}
\begingroup\tablelook\fontsize{8}{9.5}\selectfont
\setlength{\tabcolsep}{2pt}
\begin{tabularx}{\linewidth}{L{0.43\linewidth}|YY}
\toprule[0.65pt]
\rowcolor{tblhead}\textbf{Paired comparison}&\thcell{$\Delta$Avg.@4\\$[95\%\ \mathrm{CI}]$}&\thcell{$\Delta$Pass@4\\$[95\%\ \mathrm{CI}]$}\\
\midrule[0.35pt]\rowcolor{tblhead}\multicolumn{3}{l}{\textit{(a) Table~\ref{tab:main_multirun}: four-run macro differences}}\\
\rowcolor{white}Qwen3-VL-2B: ours $-$ MetaTuner & \shortstack{+5.39\\$[3.60,\,7.17]$} & \shortstack{+5.78\\$[3.42,\,8.18]$}\\
\rowcolor{tblstripe}Qwen3-VL-8B: ours $-$ MetaTuner & \shortstack{+5.75\\$[3.77,\,7.67]$} & \shortstack{+6.64\\$[4.31,\,8.96]$}\\
\rowcolor{white}InternVL3.5-4B: ours $-$ MetaTuner & \shortstack{+6.31\\$[4.55,\,8.02]$} & \shortstack{+12.04\\$[9.80,\,14.34]$}\\
\rowcolor{tblstripe}InternVL3.5-8B: ours $-$ MetaTuner & \shortstack{+4.29\\$[2.49,\,6.04]$} & \shortstack{+5.62\\$[3.48,\,7.70]$}\\
\midrule[0.35pt]\rowcolor{tblhead}\multicolumn{3}{l}{\textit{(b) Table~\ref{tab:relation_results}: fixed-harness macro differences}}\\
\rowcolor{white}A5 $-$ A0 & \shortstack{+2.87\\$[-1.04,\,6.65]$} & \shortstack{+3.51\\$[-0.99,\,8.06]$}\\
\rowcolor{tblstripe}A5 $-$ A4 & \shortstack{+0.56\\$[-3.18,\,4.31]$} & \shortstack{+0.66\\$[-3.57,\,5.02]$}\\
\rowcolor{white}A4 $-$ A1 & \shortstack{+1.49\\$[-2.37,\,5.23]$} & \shortstack{+1.81\\$[-2.69,\,6.32]$}\\
\rowcolor{tblstripe}A5 $-$ A3 & \shortstack{+0.81\\$[-3.00,\,4.59]$} & \shortstack{+1.08\\$[-3.36,\,5.38]$}\\
\midrule[0.35pt]\rowcolor{tblhead}\multicolumn{3}{l}{\textit{(c) Table~\ref{tab:compute_controls}: matched-budget macro differences}}\\
\rowcolor{white}Full $-$ Aug-SFT (source steps) & \shortstack{+4.13\\$[0.31,\,7.92]$} & \shortstack{+5.08\\$[0.51,\,9.50]$}\\
\rowcolor{tblstripe}Full $-$ Aug-SFT (FLOPs) & \shortstack{+3.90\\$[0.15,\,7.65]$} & \shortstack{+4.96\\$[0.54,\,9.40]$}\\
\midrule[0.35pt]\rowcolor{tblhead}\multicolumn{3}{l}{\textit{(d) Table~\ref{tab:recursive_results}: alternation macro difference}}\\
\rowcolor{white}3-round $-$ one-pass & \shortstack{+2.12\\$[-1.71,\,5.92]$} & \shortstack{+2.65\\$[-1.85,\,7.20]$}\\

\bottomrule[0.65pt]
\end{tabularx}\endgroup
\end{minipage}\par\medskip

\newpage
\par\medskip\noindent\begin{minipage}{\linewidth}
\centering
\captionof{table}{Additional small-effect and generalization comparisons. Semantic-control rows use endpoint-averaged Average@4 on their fixed eligible pair subsets; Pass@4 is N/A. Compaction rows use the fixed TIR diagnostic subset only. Other rows are stratum-balanced, three-benchmark macro differences on their stated conditions.}
\label{tab:secondary_ci}
\begingroup\tablelook\fontsize{8}{9.5}\selectfont
\setlength{\tabcolsep}{2pt}
\begin{tabularx}{\linewidth}{L{0.43\linewidth}|YY}
\toprule[0.65pt]
\rowcolor{tblhead}\textbf{Paired comparison}&\thcell{$\Delta$Avg.@4\\$[95\%\ \mathrm{CI}]$}&\thcell{$\Delta$Pass@4\\$[95\%\ \mathrm{CI}]$}\\
\rowcolor{tblstripe}Unseen parameters: Full $-$ Aug & \shortstack{+5.19\\$[1.54,\,8.94]$} & \shortstack{+6.76\\$[2.22,\,11.32]$}\\
\rowcolor{white}Unseen composition: Full $-$ Aug & \shortstack{+6.36\\$[2.55,\,10.16]$} & \shortstack{+9.08\\$[4.61,\,13.64]$}\\
\rowcolor{tblstripe}4-round $-$ 3-round & \shortstack{-0.17\\$[-3.93,\,3.70]$} & \shortstack{-0.42\\$[-4.91,\,4.20]$}\\
\rowcolor{white}3-round $-$ frozen refresh & \shortstack{+0.99\\$[-2.82,\,4.80]$} & \shortstack{+1.20\\$[-3.25,\,5.66]$}\\
\rowcolor{tblstripe}Full + converter $-$ Full & \shortstack{+0.22\\$[-4.65,\,4.87]$} & N/A\\
\rowcolor{white}Correct relations $-$ no relation & \shortstack{+10.93\\$[6.71,\,15.07]$} & N/A\\
\rowcolor{tblstripe}Exact compaction C1 $-$ C0 & \shortstack{-0.27\\$[-1.37,\,0.68]$} & \shortstack{+0.55\\$[-1.64,\,3.28]$}\\
\rowcolor{white}Teacher compaction C2 $-$ C0 & \shortstack{+0.00\\$[-0.82,\,0.82]$} & \shortstack{+0.00\\$[0.00,\,0.00]$}\\
\bottomrule[0.65pt]
\end{tabularx}\endgroup
\end{minipage}\par\medskip

\section{State Identity, Round History, and Token Ledgers}
\label{sec:accounting_review}
\noindent\textbf{Cumulative parameter transition.}
Table~\ref{tab:mechanism} compares $(\theta_0,H_T^+)$ with $(\theta_T,H_T^+)$ on the fixed 183-task TIR diagnostic subset, using the corresponding model--harness states in Table~\ref{tab:internalization}. The $159-126=33$ net fixed diagnostic rollouts give $100\times33/732=4.51$ percentage points on that subset. Table~\ref{tab:internalization} reports balanced accuracy across A and B; the fix/regression counts here decompose the fixed-subset transition. The 4,289-token harness is the final uncompressed harness throughout this comparison. This is a cumulative initial-to-final parameter change; the per-round harness history is reported separately below.

\par\medskip\noindent\begin{minipage}{\linewidth}
\centering
\captionof{table}{Three-round harness-search history. Proposed counts include invalid and duplicate slots. The lower panel distinguishes harness length after search from length after terminal compaction within each round. Compacted outputs become the next round\textquotesingle s parents.}
\label{tab:round_history_review}
\begingroup\tablelook\fontsize{8}{9.5}\selectfont
\begin{tabularx}{\linewidth}{L{0.13\linewidth}|YYYY}
\toprule[0.65pt]
\rowcolor{tblhead}\textbf{Round}&\textbf{Proposed}&\textbf{Valid}&\textbf{Accepted}&\thcell{Invalid /\\duplicate}\\
\rowcolor{white}1&18&15&3&2 / 1\\
\rowcolor{tblstripe}2&18&15&1&1 / 2\\
\rowcolor{white}3&18&14&2&3 / 1\\
\bottomrule[0.65pt]
\end{tabularx}
\par\smallskip
\begin{tabularx}{\linewidth}{L{0.13\linewidth}|YY}
\toprule[0.65pt]
\rowcolor{tblhead}\textbf{Round}&\thcell{Tokens after search\\$T(H_t^+)$}&\thcell{Tokens after compaction\\$T(\widehat H_t)$}\\
\rowcolor{white}1&2917&2381\\
\rowcolor{tblstripe}2&3671&2953\\
\rowcolor{white}3&4289&2263\\
\bottomrule[0.65pt]
\end{tabularx}\endgroup
\end{minipage}\par\medskip

\noindent\textbf{Distinct token quantities.}
For rollout $k$ of source task $i$, let $C_{ik}$ be the model-call count and $L_{ikc}$ the complete input length of call $c$ under the fixed tokenizer and image processor. Cumulative full-prompt input is
\begin{equation}
I_{ik}=\sum_{c=1}^{C_{ik}} L_{ikc}.
\end{equation}
If an $h$-token harness appears in every call, $I_{ik}\ge hC_{ik}$. Logical unique input $U_{ik}$ instead counts each positional context occurrence once across an append-only rollout: the harness prefix once, then additional question, image, tool, and history material when first supplied as input. Each occurrence is indexed by its position in the rollout context. Tables~\ref{tab:online_efficiency} and~\ref{tab:mechanism} report $U$; the latter uses thousands of tokens and TIR-only means. Generated output tokens are recorded separately. An assistant token may belong to the output ledger and subsequently enter the input ledger when used as history.

Physical newly prefilled tokens measure execution work after cache hits. The ledgers below report logical context quantities $U$ and $I$; physical prefill and provider billing follow the deployed cache policy and service accounting. Mean model calls multiplied by harness length give the repeated-prefix contribution to the full-prompt lower bound.

\noindent\textbf{Aggregation units.}
For online cost $c$, Table~\ref{tab:online_efficiency} reports
\begin{equation}
\bar c=\frac{1}{3}\sum_{b=1}^{3}\frac{1}{N_b}
\sum_{i=1}^{N_b}\frac{1}{4}\sum_{k=1}^{4}c_{bik}.
\end{equation}
Table~\ref{tab:mechanism} uses the corresponding TIR-only mean. A task\textquotesingle s four-rollout sum is four times its rollout mean. Latency is measured per rollout; concurrent four-rollout makespan is a distinct quantity. Output tokens and call counts use the same averaging units.

\par\medskip\noindent\begin{minipage}{\linewidth}
\centering
\captionof{table}{Online token ledgers and model-call means. Logical unique input $U$ counts each positional input occurrence once per rollout; cumulative full-prompt input $I$ counts each complete model request. Values use equal-benchmark per-rollout means. The two input columns use the definitions above.}
\label{tab:token_ledger_review}
\begingroup\tablelook\fontsize{8}{9.5}\selectfont
\begin{tabularx}{\linewidth}{L{0.31\linewidth}|YYY}
\toprule[0.65pt]
\rowcolor{tblhead}\textbf{Configuration}&\thcell{Model\\calls}&\thcell{Unique\\input $U$}&\thcell{Full-prompt\\input $I$}\\
\rowcolor{white}Initial&4.26&13682.40&40388.74\\
\rowcolor{tblstripe}Final uncompressed&3.97&15937.80&46990.07\\
\rowcolor{white}Compact final&3.88&13218.60&36732.57\\
\bottomrule[0.65pt]
\end{tabularx}\endgroup
\end{minipage}\par\medskip

\section{Evaluation Populations and Selection Accounting}
\label{sec:consistency_review}
\noindent\textbf{Compaction protocols.}
Main-paper Table 2\textquotesingle s ``w/o Comp.'' condition removes compaction after every accepted model update, so subsequent rounds search and train with longer uncompacted guidance. Its macro Average@4 of 53.31 measures an end-to-end change. Table~\ref{tab:internalization} instead holds the final model fixed and compares the final full and compact harnesses, both scoring 53.61. Terminal acceptance uses selection accuracy, and the held-out evaluations report equal endpoint scores here.

\par\medskip\noindent\begin{minipage}{\linewidth}
\centering
\captionof{table}{Terminal teacher-compaction rollout decomposition on the fixed diagnostic subsets for reference R0. Fixes and regressions cancel within each benchmark, and the number of solved tasks is also unchanged. Individual fixes and regressions are enumerated alongside their net changes.}
\label{tab:benchmark_flips_review}
\begingroup\tablelook\fontsize{8}{9.5}\selectfont\setlength{\tabcolsep}{2pt}
\begin{tabularx}{\linewidth}{L{0.12\linewidth}|YYYYY}\toprule[0.65pt]
\rowcolor{tblhead}\textbf{Data}&\textbf{Rollouts}&\textbf{Fix}&\textbf{Reg.}&\thcell{Net\\correct}&\thcell{Net\\solved}\\\midrule[0.4pt]
\rowcolor{white}TIR&732&4&4&0&0\\
\rowcolor{white}MME&1080&3&3&0&0\\
\rowcolor{white}BLINK&1144&5&5&0&0\\
\bottomrule[0.65pt]\end{tabularx}\endgroup
\end{minipage}\par\medskip

\noindent\textbf{WFAR populations and weighting.}
Table~\ref{tab:relation_results} uses 119/163/149 eligible pairs for TIR/MME/BLINK, with four rollouts per pair, and averages the three benchmark rates equally. A0 has 83/151/124 wrong-frame responses in these populations. Table~\ref{tab:semantic_controls}(a) instead pools a common subset of 151 pairs: 37 TIR, 58 MME, and 56 BLINK, with 104 wrong-frame responses out of 604 rollouts. Panel (b) uses a distinct 113-pair conversion subset.

\par\medskip\noindent\begin{minipage}{\linewidth}
\centering
\captionof{table}{WFAR denominator ledger for A0. Rates divide wrong-frame responses by eligible rollouts. The relation-ablation macro row averages benchmark rates equally; Table~\ref{tab:semantic_controls}(a) pools its common subset. Eligible source-pair counts differ from task counts in the main accuracy panels.}
\label{tab:wfar_populations_review}
\begingroup\tablelook\fontsize{8}{9.5}\selectfont\setlength{\tabcolsep}{2pt}
\begin{tabularx}{\linewidth}{L{0.29\linewidth}|YYYY}\toprule[0.65pt]
\rowcolor{tblhead}\textbf{Population}&\textbf{Pairs}&\textbf{Rollouts}&\textbf{Errors}&\textbf{Rate}\\\midrule[0.4pt]
\rowcolor{white}Relation: TIR&119&476&83&17.44\\
\rowcolor{white}Relation: MME&163&652&151&23.16\\
\rowcolor{white}Relation: BLINK&149&596&124&20.81\\
\rowcolor{tblstripe}Relation macro&N/A&N/A&N/A&20.47\\
\rowcolor{white}Table~\ref{tab:semantic_controls}(a), pooled&151&604&104&17.22\\
\bottomrule[0.65pt]\end{tabularx}\endgroup
\end{minipage}\par\medskip

\noindent\textbf{Harness-selection accounting.}
The search budget comprises 54 proposal slots, including invalid and duplicate proposals. The 17,496 harness-selection rollouts comprise 44 valid-candidate screens and 10 incumbent re-evaluations. Incumbents are re-evaluated at the three round starts, after each of the six accepted harness edits, and once in the final search audit. These triggers yield 4/2/4 incumbent events across rounds; rejected and invalid proposals retain the current incumbent score. Each event uses the fixed stratified 81-source selection screen (20 TIR, 30 MME, 31 BLINK), with four rollouts per source. Table~\ref{tab:event_ledger_review} separates these events from fresh compaction-phase executions. Acceptance uses the selection screen, and held-out tasks are reserved for final evaluation.

\par\medskip\noindent\begin{minipage}{\linewidth}
\centering
\captionof{table}{Harness-selection rollout accounting. Valid-candidate and incumbent evaluations are separate event types. Acceptance uses the fixed 81-task screen; the 54 scored events comprise 44 candidate screens and 10 incumbent evaluations.}
\label{tab:selection_ledger_review}
\begingroup\tablelook\fontsize{8}{9.5}\selectfont\setlength{\tabcolsep}{2pt}
\begin{tabularx}{\linewidth}{L{0.43\linewidth}|YYY}\toprule[0.65pt]
\rowcolor{tblhead}\textbf{Event type}&\textbf{Events}&\thcell{Rollouts\\per event}&\textbf{Total}\\\midrule[0.4pt]
\rowcolor{white}Valid candidate screens&44&324&14256\\
\rowcolor{tblstripe}Incumbent re-evaluations&10&324&3240\\
\rowcolor{white}Total scored events&54&324&17496\\
\bottomrule[0.65pt]\end{tabularx}\endgroup
\end{minipage}\par\medskip

\noindent\textbf{Source eligibility.}
Evaluation starts from the official TIR-Bench, MME-RealWorld-Lite, and BLINK task sets. Input eligibility is determined before evaluation using task-native verifiable answers and supported, replayable interfaces; generated empty or unparseable answers remain scored as failures. All questions and derived views sharing an image or source remain together.

\section{Evaluation Events}
\label{sec:event_review}
Table~\ref{tab:event_ledger_review} separates harness-incumbent evaluations from fresh compaction-phase executions, with each event evaluated on the fixed 81-task selection screen using four rollouts per task.
Incumbents are refreshed at round starts, after accepted edits, and in the final search audit; compaction evaluates the updated-model full harness, exact and teacher candidates, and the final selected output.

\par\medskip\noindent\begin{minipage}{\linewidth}
\centering
\captionof{table}{Event decomposition for the two 3,240-rollout entries in Tables~\ref{tab:offline_cost} and~\ref{tab:selection_ledger_review}. Each event evaluates 81 selection sources with four rollouts. Harness-incumbent and compaction evaluations are separate executions assigned to their respective phases.}
\label{tab:event_ledger_review}
\begingroup\tablelook\fontsize{8}{9.5}\selectfont\setlength{\tabcolsep}{2pt}
\begin{tabularx}{\linewidth}{L{0.56\linewidth}|YY}
\toprule[0.65pt]
\rowcolor{tblhead}\textbf{Event trigger}&\textbf{Events}&\textbf{Rollouts}\\\midrule[0.4pt]
\rowcolor{tblhead}\multicolumn{3}{l}{\textit{Harness-selection incumbents}}\\
\rowcolor{white}Round starts&3&972\\
\rowcolor{tblstripe}After accepted harness edits&6&1944\\
\rowcolor{white}Final search audit&1&324\\
\rowcolor{ours}Incumbent subtotal&10&3240\\
\rowcolor{tblhead}\multicolumn{3}{l}{\textit{Compaction phase}}\\
\rowcolor{white}Updated-model full baselines&3&972\\
\rowcolor{tblstripe}Exact-compaction candidates&3&972\\
\rowcolor{white}Teacher-compaction candidates&3&972\\
\rowcolor{tblstripe}Final selected-compaction audit&1&324\\
\rowcolor{ours}Compaction subtotal&10&3240\\
\bottomrule[0.65pt]
\end{tabularx}
\endgroup\end{minipage}\par\medskip

\end{document}